\pdfoutput=1

\documentclass[11pt,a4paper]{article}
\usepackage[hyperref]{style_file}
\usepackage{times}
\usepackage{hyperref}
\usepackage{latexsym}
\usepackage{microtype}
\usepackage{linguex}
\usepackage{bm}
\usepackage{tabu}
\usepackage{tabularx}
\usepackage{booktabs}
\usepackage{float}
\usepackage{multirow}
\usepackage{graphicx}
\usepackage{caption}
\usepackage{subcaption}
\usepackage{tablefootnote}
\usepackage[tone]{tipa}

\usepackage[colorinlistoftodos,prependcaption]{todonotes}
\aclfinalcopy

\title{Neural Representations for Modeling Variation in Speech}

\author{Martijn Bartelds\textsuperscript{\normalfont a} \And Wietse de Vries\textsuperscript{\normalfont a}
  \And Faraz Sanal\textsuperscript{\normalfont b} \AND Caitlin Richter\textsuperscript{\normalfont b}
  \And Mark Liberman\textsuperscript{\normalfont b} \And Martijn Wieling\textsuperscript{\normalfont ac} \AND
  \small \normalfont \textsuperscript{a}Center for Language and Cognition, Faculty of Arts, University of Groningen, Groningen, The Netherlands \\
  \small \normalfont \textsuperscript{b}Department of Linguistics, University of Pennsylvania, Philadelphia, PA, USA \\
  \small \normalfont \textsuperscript{c}Haskins Laboratories, New Haven, CT, USA \\ \AND
  \small \normalfont \textbf{Correspondence:} Martijn Bartelds, Center for Language and Cognition, Faculty of Arts, University of Groningen, \\
  \small \normalfont Oude Kijk in 't Jatstraat 26, 9712 EK Groningen, The Netherlands, E-mail: m.bartelds@rug.nl \\
  }

\date{}

\begin{document}
\maketitle
\begin{abstract}
Variation in speech is often quantified by comparing phonetic transcriptions of the same utterance. However, manually transcribing speech is time-consuming and error prone.
As an alternative, therefore, we investigate the extraction of acoustic embeddings from several self-supervised neural models. 
We use these representations to compute word-based pronunciation differences between non-native and native speakers of English, and between Norwegian dialect speakers. For comparison with several earlier studies, we evaluate how well these differences match human perception by comparing them with available human judgements of similarity. 
We show that speech representations extracted from a specific type of neural model (i.e.~Transformers) lead to a better match with human perception than two earlier approaches on the basis of phonetic transcriptions and MFCC-based acoustic features. We furthermore find that features from the neural models can generally best be extracted from one of the middle hidden layers than from the final layer. 
We also demonstrate that neural speech representations not only capture segmental differences, but also intonational and durational differences that cannot adequately be represented by a set of discrete symbols used in phonetic transcriptions.

\end{abstract}

\section{Introduction}
Past work in (e.g.,) automatic speech recognition has found that variability in speech signals is often poorly modeled, despite recent advances in speech representation learning using deep neural networks \citep{huang2014towards, huang2014comparative, koenecke2020}. This may be particularly true for monolingual as opposed to multilingual models \citep{zelasko20}.
While acoustic variability may be caused by technical aspects such as microphone variability \citep{mathur2019}, an important source of variation is the embedding of accent or dialect information in the speech signal \citep{hanani2013human, najafian2014unsupervised}.
Non-native accents are frequently observed when a second language is spoken, and are mainly caused by the first language background of non-native speakers. Similarly, regional accents are caused by the (first) dialect or regional language of the speaker.  
The accent strength of a speaker depends on the amount of transfer from their native language or dialect, and is generally influenced by a variety of characteristics, of which the age of learning the (second) language, and the duration of exposure to the (second) language are important predictors \citep{asher1969optimal, leather1983second, flege1988factors, wieling2014determinants}.

However, accent and dialect variability are often overlooked in modeling languages using speech technology, and consequently high-resource languages such as English are often treated as homogeneous \citep{blodgett2016demographic}.
Given that the number of non-native speakers of English is almost twice as large as the former group, this assumption is problematic  \citep{viglino2019end}. 
It is therefore important to accurately model pronunciation variation using representations of speech that allow accent and dialect variability to be adequately incorporated.

Traditionally, pronunciations are often represented by phonetically transcribing speech \citep{nerbonne1997measuring, livescu2000lexical, gooskens2004perceptive, heeringa2004measuring, wieling2014a, chen2016large, jeszenszky2017exploring}.
However, accurately transcribing speech using a phonetic alphabet is time consuming, labor intensive, and interference from transcriber variation might lead to inconsistencies \citep{hakkani2002active, bucholtz2007variation, novotney2010cheap}.
Additionally, phonetic transcriptions are not entirely adequate in representing how people speak, as fine-grained pronunciation differences that are relevant for studying accented speech (or dialect variation) may not be fully captured with a discrete set of symbols \citep{mermelstein1976distance, duckworth1990extensions, cucchiarini1996assessing, liberman2018}.

Consequently, acoustic-only measures have been proposed for comparing pronunciations \citep{Huckvale2007, Ferragne2010, Strycharczuk2020}. Whereas these studies only considered limited segments of speech, or exclusively included speakers from a single language background, \citet{acoustic-measure} introduced a new method that did not have these limitations. Specifically, \citet{acoustic-measure} proposed an acoustic-only method for comparing pronunciations without phonetic transcriptions, including speakers from multiple native language backgrounds while using all information available within the speech signal.
In their method, they represented accented speech as 39-dimensional Mel-frequency cepstral coefficients (MFCCs), which were used to compute acoustic-based non-native-likeness ratings between non-native and native speakers of English. 
They found a strong correlation of $r = -0.71$ between their automatically determined acoustic-based non-native-likeness scores and previously obtained native-likeness ratings provided by human raters \citep{wieling2014a}. 
This result was close to, but still not equal to, the performance of an edit distance approach on the basis of phonetic transcriptions (which showed a correlation of $r = -0.77$).

\citet{acoustic-measure} conducted several small-scale experiments to investigate whether more fine-grained characteristics of human speech were captured as compared to the phonetic transcription-based pronunciation difference measure.
Their results showed that the acoustic-only measure captured segmental differences, intonational differences, and durational differences, but that the method was not invariant to characteristics of the recording device.

The quality of MFCC representations is known to be dependent on the presence of noise \citep{zhao2013}.
Recent work has shown that neural network models for self-supervised representation learning are less affected by noise, while being well-equipped to model complex non-linear relationships \citep{schneider2019wav2vec, baevski2019vq, ling2020deep, baevski2020wav2vec}.
Generally, neural models benefit from large amounts of labeled training data.
However, self-supervised neural models learn representations of speech without the need for (manually) labeled training data.
Therefore, these models can be trained using even larger amounts of data.
Previous work has shown that fine-tuning these neural models using transcribed speech resulted in representations that resembled phonetic structure, and offered significant improvements in downstream speech recognition tasks \citep{oord2018representation, kahn2020libri}.
% Having seen large amounts of data from a large group of speakers is reason to believe that these self-supervised models may be more robust to between-speaker-within-accent variation as compared to previously used methods.
In contrast to previous methods for comparing pronunciations, these self-supervised (monolingual and multilingual) neural models are based on large amounts of data from a large group of (diverse) speakers and are therefore potentially robust against accent variation.

% Consequently, in this paper, we employ these self-supervised neural models to create a fully automatic acoustic-only pronunciation difference measure, and investigate whether this results in improved performance compared to the MFCC-based approach of \citet{acoustic-measure} and the phonetic transcription-based approach of \citet{wieling2014a}.

Consequently, in this paper, we employ and evaluate several of these self-supervised neural models in order to create a fully automatic acoustic-only pronunciation difference measure, which is able to quantify fine-grained differences between accents and dialects. Specifically, we compare and evaluate five self-supervised neural models, namely \texttt{wav2vec} \citep[subsequently denoted by \texttt{w2v}]{schneider2019wav2vec}, \texttt{vq-wav2vec} with the \texttt{BERT} extension \citep[subsequently denoted by \texttt{vqw2v}]{baevski2019vq}, \texttt{wav2vec~2.0} \citep[subsequently denoted by \texttt{w2v2}]{baevski2020wav2vec}, the multilingual \texttt{w2v2} model \texttt{XLSR-53} \citep[subsequently denoted by \texttt{XLSR}]{conneau2020unsupervised}, and \texttt{DeCoAR} \citep{ling2020deep}.
Each of these models learned speech representations by predicting short fragments of speech (e.g., approximately 300 milliseconds on average in the case of \texttt{w2v2}) within spoken sentences from the training data. These predicted fragments therefore roughly correspond to one or more subsequent phonemes (including their transitions).
These neural models were selected for this study as they achieved state-of-the-art speech recognition results on standard benchmarks such as the Wall Street Journal corpus (WSJ; \citealp{garofalo2007csr}) and the Librispeech corpus \citep{panayotov_librispeech_2015}, while differing with respect to their specific architecture.

There are several use cases in which adequately quantifying pronunciation differences automatically is important. First, the field of dialectometry (see e.g., \citealp{nerbonne1997measuring,wieling2011,wieling2015}) investigates geographical (and social) dialect variation on the basis of pronunciation differences between different dialects. While there are several dialect (atlas) datasets containing phonetic transcriptions, differences in transcription practices (sometimes even within the same dataset; \citealp{wieling2007aggregate}) limit the extent to which these pronunciations can be compared. An acoustic-only method would solve these compatibility issues, and would allow datasets that do not have phonetic transcriptions to be analyzed directly. Another use case is highlighted by a recent study of \citet{san2021leveraging}. They automatically compare pronunciations acoustically to find pronunciations of a specific word from endangered languages in a large set of unannotated speech files. Such a system, if successful, directly impacts language maintenance and revitalisation activities.

To evaluate the quality of the pronunciation differences, we will use human perceptual judgements. Previous work has shown that human listeners can adequately assess and quantify differences between pronunciations (e.g., \citealp{long1999, gooskens2005, scharenborg2007}). To determine the relative performance of our methods, we compare the use of self-supervised neural models to the phonetic-transcription-based approach of \citet{wieling2014a}, and the MFCC-based acoustic-only approach of \citet{acoustic-measure}.
More details about these methods are provided in Section~\ref{sec:methods:existing}.

To investigate the versatility and robustness of the various models, we use three different datasets for evaluation.
The first is identical to the dataset used by \citet{wieling2014a} and \citet{acoustic-measure}, and includes both acoustic recordings of native and non-native English speakers as well as human native-likeness judgements to compare against.
The second is a new dataset which focuses on accented speech from a single group of (Dutch) non-native speakers, for which human native-likeness judgements are likewise available. As we would also like to evaluate the effectiveness of the neural models for a different type of data in another language, we additionally include a dataset with Norwegian dialect pronunciations and corresponding human native-likeness ratings.
For reproducibility, we provide our code via \url{https://github.com/Bartelds/neural-acoustic-distance}.

To understand the phonetic information captured by the neural models, we introduce a visualization approach revealing the location of differences between two compared recordings, and conduct several additional small-scale experiments, in line with those conducted by \citet{acoustic-measure}.

\section{Materials}
\subsection{Datasets}
Our acoustic data comes from three datasets in two different languages.
We use two datasets that contain (mostly) non-native American-English pronunciations, and an additional dataset with Norwegian dialect pronunciations.

\subsubsection{Non-native American-English}
Pronunciations from non-native speakers of American-English are obtained from the Speech Accent Archive \citep{weinberger2015speech}, as well as the Dutch speakers dataset described in \citet{wielinglowlands}.
The Speech Accent Archive covers a wide variety of language backgrounds, while the Dutch speakers dataset is suitable for
evaluating our method on a set of English pronunciations that have more fine-grained accent differences, as it only contains speakers with the same native (Dutch) language background.

The Speech Accent Archive contains over 2000 speech samples from native and non-native speakers of English. 
Each speaker reads the same 69-word paragraph that is shown in Example~\ref{ex1}.

\ex. \label{ex1} \textit{Please call Stella. Ask her to bring these things with her from the store: Six spoons of fresh snow peas, five thick slabs of blue cheese, and maybe a snack for her brother Bob. We also need a small plastic snake and a big toy frog for the kids. She can scoop these things into three red bags, and we will go meet her Wednesday at the train station.}

Similar to past work of \citet{wieling2014a} and \citet{acoustic-measure}, we use 280 speech samples from non-native American-English speakers as our target dataset (i.e.~the non-native speakers for whom human native-likeness ratings are available), and 115 speech samples from U.S.-born L1 speakers as our reference native speaker dataset.
As there is much regional variability in the pronunciation of the native American-English pronunciations, we use a set of reference speakers (cf.~\citealt{wieling2014a}) instead of a single reference speaker. 

Among the 395 English samples from the Speech Accent Archive, 206 speakers are male and 189 speakers are female.
From these speakers, 71 male and 44 female speakers belong to the native speaker (reference) set.
The average age of the speakers in the entire dataset is 32.6 years ($\sigma=$ 13.5). Non-native speakers have an average age of onset for learning English of 10.5 years ($\sigma=$ 6.6). The 280 non-native American-English speakers have a total of 99 different native languages, with Spanish ($N = 17$), French ($N = 13$), and Arabic ($N = 12$) occurring most frequently.

The Dutch speakers dataset includes recordings of native speakers of Dutch (with no other native languages) that all read the first two sentences of the same elicitation paragraph used for the Speech Accent Archive. 
These recordings were collected at a science event held at the Dutch music festival Lowlands, where \citet{wielinglowlands} investigated the influence of alcohol on speech production in a native and non-native language. While the effect of alcohol on the pronunciation in the non-native language (English) was limited, we nevertheless only included the speech samples of all 62 sober participants (30 male and 32 female speakers). The average age of the speakers in this dataset is 33.4 years ($\sigma=$ 10.3). The average age of onset for learning English was not obtained, but generally Dutch children are exposed to English at an early age (i.e.~the subject is mandatory in primary schools from the age of about 10 to 11 onwards, but children are usually exposed to English much earlier via mass media).

For each speaker in this dataset, we phonetically transcribed the pronunciations according to the International Phonetic Alphabet. These phonetic transcriptions were created by a single transcriber (matching the conventions used by \citealt{wieling2014a}), and used to obtain the transcription-based pronunciation distances (i.e.~for comparison with the acoustic methods).

\subsubsection{Norwegian}
This dataset consists of 15 recordings and phonetic transcriptions from Norwegian dialect speakers from 15 dialect areas (4 male and 11 female speakers). The average age of these speakers is 30.5 years ($\sigma=$ 11). Moreover, each speaker lived in the place where their dialect was spoken until the mean age of 20 years, and all speakers estimated that their pronunciations were representative of the dialect they speak.

Earlier work has used this dataset for comparing dialect differences on the basis of the Levenshtein distance \citep{gooskens2004perceptive} and formant-based acoustic features \citep{heeringa2009measuring} to human perceptual dialect differences. We included this dataset and the perceptual ratings from \citet{gooskens2004perceptive} to specifically investigate whether the self-supervised neural models (even though these are, except for \texttt{XLSR}, based on the English language) are able to model differences for languages other than English. 

The speakers in this dataset all read aloud 58 words from the fable `The North Wind and the Sun'. The recordings were segmented in 58 samples corresponding to the words from the fable. For five dialects, one or two words were missing, as speakers were not always perfectly reading the text. 
Phonetic transcriptions, which we use as input for the Levenshtein distance algorithm, were created by a single transcriber.
The text, recordings, phonetic transcriptions, and transcription conventions are available online.\footnote{\url{https://www.hf.ntnu.no/nos/}.}

\subsection{Human accent and dialect difference ratings}
Human accent ratings are widely used to evaluate accentedness in speech \citep{koster1993evaluation, munro1995nonsegmental, magen1998perception, munro2001modeling}. Similarly, human ratings have been used to determine how different dialects are from each other \citep{gooskens2004perceptive}. 
To evaluate our method, we report Pearson's correlation between the computed acoustic (or phonetic transcription-based) differences and the averaged human accent (or dialect difference) ratings. 
While we evaluated read as opposed to spontaneous speech, \citet{munro1994evaluations} found that human accent ratings are not different for the two types of speech.

\subsubsection{Non-native American-English}
The perceptual data for the Speech Accent Archive speech samples were collected by \citet{wieling2014a}. 
Native U.S.-born speakers of English were invited to rate the accent strength of a set of (at most) 50 samples through an online questionnaire.
Accent strength ratings were given using a 7-point Likert scale ranging from 1 (very foreign sounding) to 7 (native English speaking abilities).
While each speech sample contained the entire 69-word paragraph (average duration of the samples was 26.2 seconds), participants were allowed to provide their rating without having listened to the full sample.
In total, the ratings of 1,143 participants were collected (57.6\% male and 42.4\% female) for a total of 286 speech samples, where each participant on average rated 41 speech samples ($\sigma=$ 14). The average amount of ratings per sample was 157 ($\sigma=$ 71). The mean age of the participants was 36.2 years ($\sigma=$ 13.9), and they most frequently lived in California (13.2\%), New York (10.1\%), and Massachusetts (5.9\%). From the 286 samples, six were from native American-English speakers. These were also identified as such, as their average ratings ranged between 6.79 and 6.97 ($0.22 \leq \sigma \leq 0.52$). 

Human accent ratings of the second (Dutch speakers) dataset were provided by a different group of U.S.-born L1 speakers of English \citep{wielinglowlands}. 
In this case, a questionnaire was created in which participants rated the accent strength of the speech samples on a 5-point Likert scale ranging from 1 (very foreign-sounding) to 5 (native English speaking abilities). Participants were not required to listen to the complete sample (average duration: 18.7 seconds) before providing their rating. A total of 115 participants (73.0\% male, 25.2\% female, and 1.8\% other) rated an average of 17 speech samples each ($\sigma=$ 9.2).
On average, each sample received 24 ratings ($\sigma=$ 6.7). The mean age of the participating raters was 47.9 years ($\sigma=$ 16). 
The participants most often originated from California (13.9\%), New York (10.4\%), and Pennsylvania (8.7\%). As the samples were shorter than for the Speech Accent Archive, a less fine-grained rating scale was used.  

The consistency of the ratings was assessed using Cronbach's alpha \citep{cronbach1951coefficient}. For both studies, the ratings were consistent, with alpha values of 0.85 and 0.92 for the Speech Accent Archive dataset and Dutch speakers dataset, respectively \citep{nunnally1978}.

\subsubsection{Norwegian}
\citet{gooskens2004perceptive} carried out a listening experiment using the recordings of the Norwegian dataset. A total of 15 groups of raters (high school pupils, one group per dialect area) were asked to judge each speaker on a 10-point scale. A score of 1 was given when the pronunciation of the speaker was perceived to be similar to the rater's own dialect, while a score of 10 indicated that the pronunciation of the speaker was maximally dissimilar from the rater's own dialect. The average duration of the speech samples was about 31 seconds.

On average, each group consisted of 19 listeners (48\% male and 52\% female) with a mean age of 17.8 years. For the majority of their life (16.7 years, on average), raters had lived in the place where their dialect was spoken. Only 3\% of the raters reported to never speak in their local dialect. About 81\% of the raters reported to use their dialect often or always. The consistency of the ratings was not reported by \citet{gooskens2004perceptive}. 

\section{Methods}
\label{sec:methods}
\subsection{Self-supervised neural models}
\label{sec:models}
% We compare and evaluate five self-supervised pre-trained neural models, namely \texttt{wav2vec} \citep[subsequently denoted by \texttt{w2v}]{schneider2019wav2vec}, \texttt{vq-wav2vec} \citep[subsequently denoted by \texttt{vqw2v}]{baevski2019vq}, \texttt{wav2vec~2.0} \citep[subsequently denoted by \texttt{w2v2}]{baevski2020wav2vec}, the multilingual \texttt{w2v2} model \texttt{XLSR-53} \citep[subsequently denoted by \texttt{XLSR}]{conneau2020unsupervised}, and \texttt{DeCoAR} \citep{ling2020deep}.

%Regarding the \texttt{XLSR} model, we exclusively use models which have been fined-tuned on a language from the Common Voice dataset suitable for our individual datasets \citep{ardila2019common}.
%Specifically, we apply \texttt{XLSR-en} (fine-tuned on 1,686 hours of English) to the non-native English speech datasets and \texttt{XLSR-sv} (fine-tuned on 12 hours of Swedish, which was the most similar available language to Norwegian) to the Norwegian dialects dataset.

We compare and evaluate five self-supervised pre-trained neural models (i.e.~\texttt{w2v}, \texttt{vqw2v}, \texttt{w2v2}, \texttt{XLSR}, and \texttt{DeCoAR}).
The self-supervised neural models have learned representations of acoustic recordings by training the models to predict upcoming speech frames, without using labeled data \citep{schneider2019wav2vec, ling2020deep, baevski2019vq, baevski2020wav2vec}. An important characteristic of these deep learning models is that they contain multiple hidden layers containing information about the underlying data.  
Architectures and training techniques of these models have typically been inspired by successful methods in natural language processing such as \texttt{word2vec} \citep{mikolov_distributed_2013}, \texttt{ELMo} \citep{petersDeepContextualizedWord2018}, and \texttt{BERT} \citep{devlin_bert_2019}.

% In short, \texttt{w2v} uses a multi-layer convolutional neural network to learn speech representations from raw audio.
% \texttt{vqw2v} extends \texttt{w2v} by creating discrete token representations based on a \texttt{w2v} encoder.
% The discrete token representations are subsequently used as input to a 12-layer \texttt{BERT} model, which is trained separately.
% In contrast to \texttt{vqw2v}, the representations of \texttt{w2v2} are learned end-to end.
% While \texttt{w2v2} consists of an encoder, a quantizer, and a 24-layer Transformer network, the encoder representations are optimized for use in the Transformer network.
% The architecture of \texttt{XLSR} is similar to \texttt{w2v2}, except that the model learns a single set of quantized speech representations based on the encoder representations. Sharing this set across pre-training languages was found to be successful for learning multilingual speech representations.
% Finally, \texttt{DeCoAR} uses a stacked bi-directional long short-term memory (LSTM) network for learning speech representations.

All of the evaluated acoustic models, except \texttt{XLSR}, were pre-trained on the large unlabeled Librispeech dataset, which contains 960 hours of English speech obtained from audio books (LS960). This dataset is divided into two parts, namely a part which includes clean data (460 hours), and a part which includes noisy data (500 hours).
Speakers with accents closest to American-English (represented by pronunciations from the Wall Street Journal-based CSR corpus (SI-84) described by \citealt{paul-baker-1992-design}) were included in the clean data part, while the noisy data part contained accents that were more distant from American-English \citep{panayotov_librispeech_2015}.
The \texttt{XLSR} model, instead, was trained on 56,000 hours of speech from a total of 53 languages, including European, Asian, and African languages.
Note that the majority of the pre-training data for \texttt{XLSR} still consists of English speech (44,000 hours).

In addition to the pre-trained model variants, there are fine-tuned variants available for the \texttt{w2v2} and \texttt{XLSR} models.
These models were fine-tuned on labeled data in a specific language to improve their performance on speech recognition tasks.
However, the process of fine-tuning might have influenced the linguistic representations that are learned during pre-training.
We therefore also include these fine-tuned model variants in our evaluation.
For English, we evaluate the \texttt{w2v2} model that has been fine-tuned on 960 hours of English speech data (subsequently denoted by \texttt{w2v2-en}), and the \texttt{XLSR} model that was fine-tuned on 1,686 hours of English speech data (further denoted by \texttt{XLSR-en}). The \texttt{w2v2-en} model was chosen because it is the largest fine-tuned English model available, and \citet{baevski2020wav2vec} showed that increasing the model size improved performance on all evaluated speech recognition tasks. For Norwegian, we included the \texttt{XLSR} model fine-tuned on 12 hours of Swedish (which was the closest language available to Norwegian with a fine-tuned model available; further denoted by \texttt{XLSR-sv}).
% The (technical) details of these models and their variants can be found in Appendix~\ref{appendix:models}. 

The effectiveness of these self-supervised neural models was originally evaluated by using the learned representations for the task of automatic speech recognition.
However, in this study we assess whether or not these acoustic models also capture fine-grained information such as pronunciation variation. 
% As several of the investigated algorithms (specifically, \texttt{vqw2v}, \texttt{w2v2}, and \texttt{XLSR}) use a Transformer network to model the acoustic signal, we also evaluate (using a development set) which Transformer layers are most suitable for our task.
As the investigated algorithms use multiple hidden layers to model the acoustic signal, we also evaluate (using a development set) which layers are most suitable for our specific task.
More information about these and other aspects of the models can be found in Appendix~\ref{appendix:models} and \ref{appendix:layers}.

\subsection{Existing methods}
\label{sec:methods:existing}
For comparison with the self-supervised neural models, we also report the results on the basis of two existing approaches for quantifying pronunciation differences, namely the MFCC-based approach of \citet{acoustic-measure} and the phonetic transcription-based approach of \citet{wieling2012inducing}. Both methods are currently the best-performing automatic (acoustic- or transcription-based) algorithms for determining pronunciation differences that match human perceptual pronunciation differences well, and are explained in more detail below.

\subsubsection{Phonetic transcription-based distance calculation}
The phonetic transcription-based distances are determined on the basis of the adjusted Levenshtein distance algorithm proposed by \citet{wieling2012inducing}. The Levenshtein algorithm determines the cost of changing one phonetically transcribed pronunciation into another by counting the minimum amount of insertions, deletions, and substitutions \citep{levenshtein}. The adjustment proposed by \citet{wieling2012inducing} extends the standard Levenshtein distance by incorporating sensitive segment differences (rather than the binary distinction of same vs.~different) based on pointwise mutual information (PMI) \citep{church1990word}. This data-driven method assigns lower costs to sound segments that frequently occur together, while higher costs are assigned to pairs of segments that occur infrequently together. These sensitive sound segment differences are subsequently incorporated in the Levenshtein distance algorithm. An example of a PMI-based Levenshtein alignment for two pronunciations of the word ``afternoon'' is shown in Figure~\ref{fig:PMI-LD}. 

\begin{figure}[ht]
    \begin{center}
        \begin{tabular}{@{}ccccccccc@{}}
            \textipa{\ae} & \textipa{@} & \textipa{f} & \textipa{t} & \textipa{@} & \textipa{} & \textipa{n} & \textipa{\textbaru} & \textipa{n} \\
            \textipa{\ae} & \textipa{} & \textipa{f} & \textipa{t} & \textipa{@} & \textipa{r} & \textipa{n} & \textipa{u} & \textipa{n} \\
            \hline
             & .031 &  & &  & .030 & & .020 &
        \end{tabular}
    \caption{PMI-based Levenshtein alignment for two different pronunciations of the word ``afternoon''. The total transcription-based pronunciation distance between the two pronunciations equals the sum of the costs of all edit operations (i.e.~0.081).}
    \label{fig:PMI-LD}
    \end{center}
\end{figure}

To obtain reliable segment distances using the PMI-based Levenshtein distance algorithm, it is beneficial if the number of words and segments is as large as possible. As the Dutch speakers dataset is relatively small, we instead used the sensitive segment differences obtained on the basis of the (larger) Speech Accent Archive dataset (i.e.~the same as those used by \citealp{wieling2014a}).

After the Levenshtein distance algorithm (incorporating sensitive sound differences) is used to quantify the pronunciation difference between each word for a pair of speakers, the pronunciation difference between two speakers is subsequently determined by averaging all word-based pronunciation differences. Additionally, for the two English datasets, the difference between the pronunciation of a non-native speaker and native (American-English) speech (i.e.~the non-native-likeness) is computed by averaging the pronunciation difference between the non-native speaker and a large set of native English speakers (the same for both datasets).

\subsubsection{MFCC-based acoustic distance calculation}
For the Speech Accent Archive recordings, the MFCC-based differences between the individual non-native speakers and native English speakers were available from \citet{acoustic-measure}. 
For the native Dutch speakers dataset, and the Norwegian dataset, we calculate these differences following the same approach. 
In short, this consists of comparing 39-dimensional MFCCs of pronunciations of the same word (by two speakers) to obtain the acoustic difference between the pronunciations.
We use dynamic time warping to compare the MFCCs \citep{giorgino2009}.
This algorithm is widely used to compare sequences of speech features by computing the minimum cumulative distance (i.e.~the shortest path) through a cost matrix that contains the Euclidean distance between every pair of points in the feature representations.
To account for durational differences between the pronunciations, we normalize the minimum cumulative distance by the length of the feature representations.
See \citet{acoustic-measure} for more details. Finally, the non-native-likeness is computed in the same way as for the Levenshtein distance algorithm, explained in the previous section. 
%The pronunciation difference between two speakers is subsequently determined by averaging all word-based pronunciation differences.
%Finally, the difference between the pronunciation of a non-native speaker and native (American-English) speech (i.e. the non-native-likeness) is computed by averaging the pronunciation difference between the non-native speaker and a large set of native speakers.

\section{Experimental setup}

\subsection{Non-native American-English pronunciation differences}
Following \citet{wieling2014a} and \citet{acoustic-measure}, we compute a measure of acoustic distance from native English speech by individually comparing the non-native target samples from both datasets to the 115 native reference samples. 
Neural representations of all samples are acquired by using the full samples as input to the neural models.
The final output of these neural models should correspond with the original input (including all frames), and will therefore not contain any new information.
Because of this, we use the feature representations of hidden layers (discussed in Section~\ref{sec:models}) as acoustic embeddings.
These representations are extracted by doing a forward pass through the model up to the target hidden layer.
% For the Transformer-based methods (\texttt{vqw2v}, \texttt{w2v2}, and \texttt{XLSR}), it is in principle possible to choose which layer is selected as the output layer.
Specifically, we investigated for each neural model which layer performed best for our task, by evaluating the performance (i.e.~the correlation with human ratings) using a held-out development set (25\% of the data of the Speech Accent Archive dataset, and 50\% of the data of the much smaller Dutch speaker dataset). As layers sometimes show very similar performance, we also evaluated which layers showed significant lower performance than the best-performing layer. For this, we used the modified $z$-statistic of \citet{steiger1980tests} for comparing dependent correlations. After selecting the best-performing layer, the performance is evaluated on the remaining data (and the full dataset, if the patterns of the development set and the other data are similar). Samples are cut into individual words after embedding extraction using time-alignments from the Penn Phonetics Lab Forced Aligner \citep{yuan2008speaker}.
For word pairs between a reference and target speaker, length normalized similarity scores between the embeddings are calculated using dynamic time warping.

Scores are averaged across all 69 words (Speech Accent Archive dataset) or 34 words (Dutch speakers dataset) to acquire a distance measurement between a target speaker and a reference speaker. To compute a single score of distance between a target speaker and native English speech, the distances between the target speaker and all reference native speakers are averaged. 

%The MFCC-based acoustic method and the phonetic transcription-based method follow the same experimental setup to compute pronunciation differences (i.e. calculating word-based differences per speaker pair, averaging over all words, and comparing each non-native speaker to the full reference native speaker set). 

\begin{figure*}[t]
  \begin{center}
    \includegraphics[width=\textwidth]{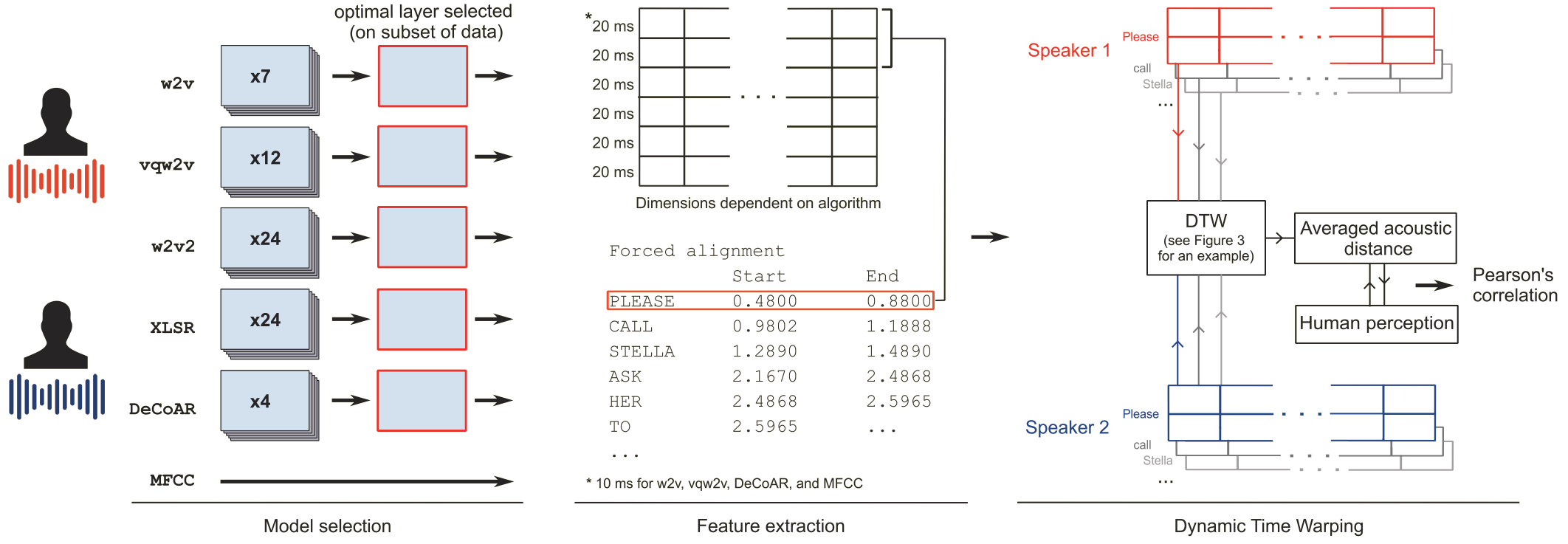}
    \caption{Visualization of the acoustic distance measure where features are extracted using several acoustic-only methods. The output layer of the models is selected in a validation step. After feature extraction, the samples are sliced into individual words, which are subsequently compared using dynamic time warping. The word-based acoustic distances are then averaged and compared to human perception.}
    \label{fig:measure}
  \end{center} 
\end{figure*}

We evaluate our algorithms on both datasets by calculating the Pearson correlation between the resulting acoustic distances and the averaged human native-likeness judgements for the target samples. Note, however, that the results on the basis of the Speech Accent Archive are likely more robust as this dataset contains a large amount of (longer) samples, a variety of native language backgrounds, and a larger amount of ratings per sample.
We visualize the complete approach in Figure~\ref{fig:measure}.

\subsection{Norwegian pronunciation differences}
For the Norwegian dataset, we measure acoustic distances by computing neural representations for the segmented word samples similar to the approach used for the non-native American-English samples. 
% The selection of the best-performing layer for the Transformer-based methods was determined again using a validation set, containing a random sample of 50\% of the data.
The selection of the best-performing layer for the neural methods was determined again using a validation set, containing a random sample of 50\% of the data.
%For the Dutch dialects dataset, we did not have human perceptual judgements available, but there we chose the layer with the highest correlation with the phonetic transcription-based distance. 
Word-based neural representations of the same word are compared using dynamic time warping to obtain similarity scores, which are length normalized.
These are subsequently averaged to obtain a single distance measure between two dialects (i.e.~two speakers).

We evaluate our algorithms on the Norwegian dialects dataset by computing the Pearson correlation between the acoustic distances and perception scores provided by the dialect speakers, and compare this value to the correlation obtained by using phonetic transcription-based distances and MFCC-based distances instead of the self-supervised neural acoustic-only distances. As \citet{gooskens2004perceptive} found that dialect distances with respect to themselves erroneously increased the correlation with the perceptual distances, we excluded these distances from our analysis.

% For the Dutch dialects dataset, unfortunately, we have no perceptual distances available. We therefore evaluate our algorithms by computing the Pearson correlation between the acoustic-only distances and phonetic transcription-based distances of \citet{wieling2007aggregate} using the same set of words for matching localities. Furthermore, we create multidimensional scaling (MDS; \citealp{torgerson1952multidimensional}) maps of the different algorithms visualizing the relationship between the location of each speaker and the pronunciation differences. These maps can be compared qualitatively, by assessing whether traditional dialect differences are visible.

\subsection{Influence of sample}
To obtain a better understanding of the influence of our reference sample, and the specific set of words on our results, we conduct several additional experiments on the (larger) Speech Accent Archive non-native English dataset using our best-performing model.

First, we investigate the effect of choosing a single reference speaker, as opposed to using the reference set of all 115 speakers. Second, we further examine the effect of speaker backgrounds on the correlation with human perception, by restricting the set of reference native speakers to speakers from the western half of the U.S.~and the English-speaking part of Canada. We opt for this set, as these areas are characterized by less dialect variation compared to the eastern half of the U.S.~\citep{boberg2010english}. Third, as the gender distribution between the native and non-native speakers differed for our reference speaker set compared to the set of non-native speakers, we investigate the influence of gender by restricting the reference set to a single gender.

Finally, while the correlations are determined on the basis of an average over 69 words, we are also interested in the performance when only individual words are selected. This analysis may reveal which words are particularly informative when determining non-native-likeness.

\subsection{Understanding representations}
To obtain a better understanding of the acoustic properties to which our final best-performing neural acoustic distance measure is sensitive, we conduct several additional experiments using the Speech Accent Archive recordings. We first evaluate how well the models are able to capture variation in specific groups of non-native speakers. By restricting the background (i.e.~the native language) and thereby creating a more homogeneous sample (similar to the Dutch speakers dataset), human accent ratings may lie closer together. Strong correlations between human perception and acoustic distances when the range of scores is large (as in the full dataset), may not necessarily also imply strong correlations when there is less variation. Consequently, this experiment, together with the analysis of the Dutch speakers data, investigates whether or not our models also model human perception at a more fine-grained level. 

In addition, to understand whether the acoustic distances comprise (linguistically relevant) aspects of pronunciation different from pronunciation distances computed using MFCCs or phonetic transcriptions, we fit multiple linear regression models. In those models, human accent ratings are predicted based on the acoustic distances of our best-performing self-supervised neural model, MFCC-based acoustic distances \citep{acoustic-measure}, and phonetic transcription-based differences \citep{wieling2014a}. We evaluate the contribution of each predictor to the model fit, and assess the model's explained variance to determine whether distinctive aspects of pronunciation are captured.

% Finally, \citet{acoustic-measure} found that acoustic distances computed by using MFCCs cannot only capture segmental differences, but also intonational and durational differences. To assess if this information is embedded in our method as well, and to conduct a fair comparison, we include two of the experiments conducted by \citet{acoustic-measure}. First, we compute acoustic distances between ten repetitions of hVd-words using 12 Dutch monophthongs (\textipa{a, A, E, e, \o, I, i, O, u, o, Y, y}) where all recordings are pronounced by the same speaker. These distances are then correlated with Bark-scaled formant distances based on the frequencies of the first and second formant measured at each mid-point of the vowel. Both the formant-based distances and the acoustic-based distances are also visualized using multidimensional scaling \citep{torgerson1952multidimensional}. Next, we assess whether our method can distinguish aspects of speech that are linguistically relevant from those which are not. We therefore compute acoustic distances between four series of recordings of the word ``living'' (ten repetitions per series) and compare the acoustic distances to those computed using MFCCs. The first two series of recordings were unmodified but recorded with a different recording device (the built-in microphone of a laptop, versus the built-in microphone of a smartphone). The third and fourth series were manipulated by changing the intonation (``living?'') and relative duration of the first syllable (``li\_ving''), respectively.

Finally, \citet{acoustic-measure} found that acoustic distances computed by using MFCCs not only captured segmental differences, but also intonational and durational differences between acoustically altered pronunciations of the same word. To assess whether this information is captured by our best-performing neural method as well, we replicate the experiment of \citet{acoustic-measure}. Specifically, we compute acoustic distances between four series of recordings of the word ``living'' (ten repetitions per series) and compare the acoustic distances to those computed using MFCCs. The first two series of recordings were unmodified but recorded with a different recording device (the built-in microphone of a laptop, versus the built-in microphone of a smartphone). The third and fourth series were manipulated by changing the intonation (``living?'') and relative duration of the first syllable (``li:ving''), respectively. To illustrate the results of this experiment, we have developed a visualization tool, which is discussed below and may help understand whether or not our best-performing (black box) neural method is able to distinguish aspects of speech that are linguistically relevant from those that are not.

\subsubsection{Visualization tool}
% Actual files:
% falling_huud_mobiel_201145.wav
% falling_hood_mobiel_203936.wav

\begin{figure}[ht]
  \begin{center}
    \includegraphics[width=3in]{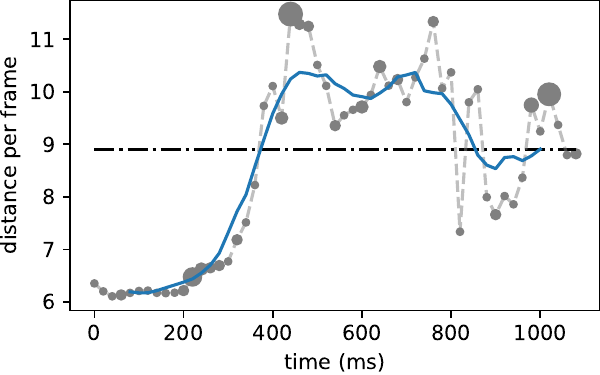}
    \caption{Visualization of neural acoustic distances per frame (based on \texttt{w2v2}) with the pronunciation of /\textipa{hy:d}/ on the $x$-axis and distances to the pronunciation of /\textipa{ho:d}/ on the $y$-axis. The horizontal line represents the global distance value (i.e.~the average of all individual frames). The blue continuous line represents the moving average distance based on 9 frames, corresponding to 180ms. As a result of the moving average, the blue line does not cover the entire duration of the sample. Larger bullet sizes indicate that multiple frames in /\textipa{ho:d}/ are aligned to a single frame in /\textipa{hy:d}/. }
    \label{fig:hXXd-distance}
  \end{center} 
\end{figure}

For this study, we have developed a tool that visualizes the dynamic time warping alignments and the corresponding alignment costs to highlight where in the acoustic signal the differences between two pronunciations of the same word is most pronounced. As such, this tool may be helpful for interpreting the acoustic distances returned by our models, for example by highlighting that the acoustic differences between two pronunciations are most divergent at the end (or start) of a word. An illustration of the output (and interpretation) of this tool is shown in Figure~\ref{fig:hXXd-distance}, which compares the pronunciation of a Dutch speaker pronouncing the two non-words /\textipa{hy:d}/ vs.~/\textipa{ho:d}/. This example illustrates the relative influence of different phonemes on the acoustic distance within a word. The difference between the two pronunciations is lowest in the beginning of the word (/\textipa{h}/), whereas it is highest in the middle part (comparing [\textipa{y:}] and [\textipa{o:}]). The difference at the end (i.e.~/\textipa{d}/) is higher than at the beginning (for /\textipa{h}/), which may reflect perseverative coarticulation, despite the transcriptions being identical. An online demo of this visualization tool can be used to generate similar figures for any pair of recorded pronunciations.\footnote{\url{https://bit.ly/visualization-tool}}

\section{Results}
We first report on the performance of the non-native American-English speakers from the Speech Accent Archive and Dutch speakers dataset.
Subsequently, we present the results on the Norwegian dataset to show how the self-supervised models perform on a language different from English.
Finally, we discuss the phonetic information encoded in the pre-trained representations using visualizations of the acoustic distances, and report on the results from our additional experiments.

%\subsection{Transformer layer performance}
% \paragraph{Transformer layer performance} 
%To investigate the information captured by the Transformer layers (\texttt{BERT} extension to \texttt{vqw2v}, \texttt{w2v2}, \texttt{XLSR}, \texttt{w2v2-en}, and \texttt{XLSR-en}), we determine the best-performing output layer on our task.
%For that purpose, we compute Pearson correlations between acoustic distances (computed using the different Transformer layers) and human native-likeness ratings on a 25\% subset (i.e.~a validation set) from the Speech Accent Archive dataset and 50\% subset from Dutch speaker dataset.
%The latter validation set is larger, as preliminary results showed that a 25\% subset was insufficient to yield a performant validation set.
%We moreover indicate whether the differences between the layers are significant using the modified \textit{z}-statistic of \citet{steiger1980tests} for comparing dependent correlations.
%The performance of the optimal output layers are subsequently evaluated on the full dataset excluding the validation subset (i.e.~a test set), ensuring that our results are generalizable.
%The direction of the correlations is negative, as the acoustic distances measure pronunciation difference from native English, while the human perceptual ratings indicate similarity with native English pronunciations.

\subsection{Non-native American-English pronunciation differences}

Table~\ref{table:cors} shows the correlations between the non-native-likeness scores and the average human native-likeness ratings for both datasets. The modified \textit{z}-statistic of \citet{steiger1980tests} shows that the \texttt{w2v2-en} model significantly outperforms all other models (including the Levenshtein distance approach, which was already reported to match human perception well; \citealp{wieling2014a}) when applied to the Speech Accent Archive dataset (all $z\textrm{'s} > 3$, all $p\textrm{'s} < 0.001$). Similarly, for the Dutch speakers dataset, the \texttt{w2v2-en} is also the best-performing model. In this case, it significantly improved over \texttt{w2v}, \texttt{vqw2v}, \texttt{DeCoAR}, \texttt{XLSR}, and \texttt{MFCC} (all $z\textrm{'s} > 3$, all $p\textrm{'s} < 0.001$), but not over the other approaches ($p > 0.05$). 

\begin{table}[ht!]
\centering
    \begin{tabular}{lrr}
        \toprule
        \textbf{Model} & \textbf{SAA} & \textbf{DSD}\\
        \midrule
        \texttt{w2v} (7, 5) & -0.69 & -0.25\\
        \texttt{vqw2v} (11, 10) & -0.78 & -0.67\\
        \texttt{w2v2} (17, 12) & -0.85 & -0.70\\
        \texttt{XLSR}\tablefootnote{\label{f1}We also computed correlation coefficients using the most recent \texttt{XLS-R} model \citep{babu2021xlsr}, which is pre-trained on 436,000 hours of speech in 128 languages. To directly compare the results to \texttt{XLSR} and \texttt{XLSR-en}, we used the pre-trained model with the same number of parameters and fine-tuned this model on English labeled data available in the Common Voice dataset. However, the results of these newer models are not significantly better ($p > 0.05$) from the results obtained using \texttt{XLSR} and \texttt{XLSR-en}. We therefore report those latter results.} (16, 16) & -0.81 & -0.47\\
        \texttt{DeCoAR} (2, 4)  & -0.62 & -0.40\\
        \midrule
        \texttt{w2v2-en} (10, 9) & \textbf{-0.87} & \textbf{-0.71}\\
        \texttt{XLSR-en}\textsuperscript{\getrefnumber{f1}} (8, 9) & -0.81 & -0.63\\
        \midrule
        LD \citep{wieling2014a} & -0.77 & -0.70\\
        MFCC \citep{acoustic-measure} & -0.71 & -0.34\\
        \bottomrule
    \end{tabular}
\caption{Pearson correlation coefficients $r$ between acoustic-only or phonetic transcription-based distances and human native-likeness ratings, using \texttt{w2v}, \texttt{vqw2v}, \texttt{w2v2}, \texttt{XLSR}, \texttt{w2v2-en}, \texttt{XLSR-en}, \texttt{DeCoAR}, the PMI-based Levenshtein distance (LD), and MFCCs to compute distances on the Speech Accent Archive (SAA) dataset and native Dutch speakers dataset (DSD). All correlations are significant at the $p < 0.001$ level. The values between parentheses show the selected layers of the neural models on the basis of the 25\% validation set for the Speech Accent Archive dataset and the 50\% validation set for the Dutch speakers dataset, respectively.}
\label{table:cors}
\end{table}

\begin{figure}[ht]
  \begin{center}
    \includegraphics[width=3in]{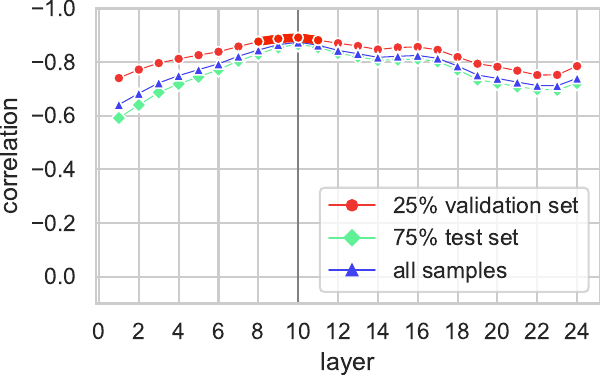}
    \caption{Pearson correlation coefficients of acoustic distances compared to human accent ratings for different Transformer layers in the \texttt{w2v2-en} model. The vertical line marks the layer that was chosen as the best-performing layer based on the 25\% validation set of the Speech Accent Archive dataset. Layers with a correlation that is  not significantly different ($p > 0.05$) from the optimal layer are indicated by the thick red line.}
    \label{fig:wav2vecen}
  \end{center} 
\end{figure}

% For the Transformer-based models, the numbers between parentheses show the best-performing layer (on the basis of the performance on the validation set).
For the neural models, the numbers between parentheses show the best-performing layer (on the basis of the performance on the validation set).
As an example of how individual layers may show a different performance, Figure~\ref{fig:wav2vecen} shows the performance for each layer for the best-performing \texttt{w2v2-en} model applied to the Speech Accent Archive dataset. It is clear that rather than selecting the final layer, the performance of an intermediate layer (10) is highest (and not significantly different from the performance of layers 8 to 11). Furthermore, there is a close match between the observed pattern for both the validation set and the test set. 
% Appendix~\ref{appendix:layers} shows these graphs for all Transformed-based algorithms and datasets. 
Appendix~\ref{appendix:layers} shows these graphs for all neural models and datasets. 

\subsection{Norwegian pronunciation differences}
% In this section, we only consider neural representations based on the fine-tuned model variants of \texttt{w2v2} and \texttt{XLSR}, as these models showed the best performance compared to their respective non-fine-tuned models in the previous section. Specifically, we evaluate \texttt{w2v2-en} and \texttt{XLSR-sv} for the Norwegian dialects dataset, and \texttt{w2v2-en} and \texttt{XLSR-nl} for the Dutch dialects dataset. 

%The direction of the correlations for the Norwegian dataset is no longer negative, as the acoustic distances and human perceptual ratings both measure pronunciation differences between the Norwegian dialects.
% Also note that we excluded acoustic distances and perceptual ratings from dialects with respect to themselves when computing the correlations, because they mistakenly increased the correlation scores (see \citealt{gooskens2004perceptive}).

% \subsubsection{Norwegian dialects dataset}

Table~\ref{table:cors-NOS} shows the results for the Norwegian dialects dataset.
% We only consider neural representations based on the fine-tuned model variants of \texttt{w2v2} and \texttt{XLSR}, as these models showed the best performance compared to their respective non-fine-tuned models in the previous section.
In this experiment, we only include neural representations from the best-performing fine-tuned monolingual English and multilingual model in the previous section (i.e.~\texttt{w2v2-en} and \texttt{XLSR-sv} as Swedish is more similar to Norwegian than English). Unfortunately, there is no monolingual Norwegian model available.
In this case, the performance of the PMI-based Levenshtein distance is substantially (and significantly: all $z\textrm{'s} > 3$, all $p\textrm{'s} < 0.001$) higher than both of the neural methods (which did not differ from each other in terms of performance, but were  improve over the MFCC approach; $z > 3$, $p < 0.001$). Note that the correlations are positive, as higher perceptual ratings reflected more different dialects.

\begin{table}[ht!]
\centering
    \begin{tabular}{lr}
        \toprule
        \textbf{Model} & \textbf{Mean \textit{r}}\\
        \midrule
        \texttt{w2v2-en} (3) & 0.49 \\
        \texttt{XLSR-sv}\tablefootnote{When using \texttt{XLS-R} fine-tuned on Swedish labeled data from the Common Voice dataset, the correlation coefficient is not significantly different ($p > 0.05$) from \texttt{XLSR-sv}.} (7) & 0.49 \\
        \midrule
        LD \citep{wieling2014a} & \textbf{0.66} \\
        MFCC \citep{acoustic-measure} & 0.22 \\
        \bottomrule
    \end{tabular}
\caption{Pearson correlation coefficients $r$ between acoustic-only or phonetic transcription-based distances and human native-likeness ratings, using \texttt{w2v2-en}, \texttt{XLSR-sv}, the PMI-based Levenshtein distance (LD), and MFCCs for computing pronunciation distances for the Norwegian dialect dataset. All correlations are significant at the $p < 0.001$ level. The values between parentheses show the selected layers of the neural models on the basis of the 50\% validation set.}
\label{table:cors-NOS}
\end{table}

\subsection{Influence of sample}
% Individual speaker variation
% Ref spk from west US canada
% Gender differences
% cor per word
In this section, we report on the influence of the specific sample of reference speakers and the included words across which we averaged.
Table~\ref{table:singleref} reveals the influence of our specific sample of reference speakers by showing the averaged correlation coefficients (and the associated standard deviation) for the various methods applied to the Speech Accent Archive dataset. Instead of using the full set of 115 native speakers as reference set, in this analysis each individual native speaker was used once as the single reference speaker. Particularly of note is that only \texttt{w2v2}, \texttt{XLSR} and their fine-tuned variants, as well as the PMI-based Levenshtein distance appear to be minimally influenced by individual reference speaker differences (i.e.~reflected by the low standard deviations).
Specifically, \texttt{w2v2} and \texttt{w2v2-en} yield the lowest standard deviations as well as the highest correlation ranges for individual reference speakers.

%both fine-tuned models, as well as 
%We observe that the representations of \texttt{w2v2} and \texttt{w2v2-en} are minimally affected by individual speaker differences, while the other models show substantially decreased correlations or a much larger spread depending on the individual reference speakers compared to the results discussed previously.
%Interestingly, we find that \texttt{w2v2}, \texttt{w2v2-en}, and \texttt{XLSR-en} not only perform better than the acoustic-only methods in this comparison, but also compared to the phonetic transcription-based distances computed using the PMI-based Levenshtein distance.
%Particularly striking is the difference in correlation range between either \texttt{w2v2} or \texttt{w2v2-en} and the PMI-based Levenshtein distance.
%This suggests a greater stability of \texttt{w2v2} and \texttt{w2v2-en} with respect to individual speaker variation.

\begin{table*}[ht!]
\centering
    \begin{tabular}{lrrr}
      \toprule
      \textbf{Model} &\textbf{Mean \textit{r}} & \textbf{Std. Dev.} & \textbf{Range}\\
      \midrule
      \texttt{w2v} (7) & -0.57 & 0.11 & [-0.14, -0.73]\\
      \texttt{vqw2v} (11) & -0.69 & 0.08 & [-0.16, -0.79]\\
      \texttt{w2v2} (17) & -0.83 & 0.02 & [-0.73, -0.86]\\
      \texttt{XLSR} (16) & -0.76 & 0.05 & [-0.47, -0.83]\\
      \texttt{DeCoAR} (2) & -0.49 & 0.08 & [-0.22, -0.67]\\
      \midrule
      \texttt{w2v2-en} (10) & \textbf{-0.86} & \textbf{0.01} & \textbf{[-0.79, -0.88]}\\
      \texttt{XLSR-en} (8) & -0.78 & 0.04 & [-0.53, -0.83]\\
      \midrule
      LD \citep{wieling2014a} & -0.74 & 0.04 & [-0.52, -0.79]\\
      MFCC \citep{acoustic-measure} & -0.45 & 0.10 & [-0.20, -0.69]\\
      \bottomrule
    \end{tabular}
\caption{Averaged Pearson correlation coefficients $r$, with standard deviations and correlation ranges, between acoustic-only or phonetic transcription-based distances and human native-likeness ratings applied to the Speech Accent Archive dataset, using \texttt{w2v}, \texttt{vqw2v}, \texttt{w2v2} (pre-trained and fine-tuned), \texttt{XLSR} (pre-trained and fine-tuned), \texttt{DeCoAR}, the PMI-based Levenshtein distance (LD), and MFCCs to compute distances when individual U.S.-born native American-English speakers were treated as the single reference speaker. All correlation coefficients are significant at the $p < 0.001$ level. The values between parentheses show the selected layer of the neural models on the basis of the validation set.}
\label{table:singleref}
\end{table*}

Additionally, we computed the correlation coefficient using our best-performing model (i.e.~\texttt{w2v2-en}) based solely on including reference native speakers from the western half of the U.S.~and the English-speaking part of Canada. The resulting correlation of $r = -0.87$ ($p < 0.001$) was identical to the correlation when including all reference speakers. The results were also similar when the reference speaker set was restricted to only men or women, with correlations of $r = -0.87$ ($p < 0.001)$ and $r = -0.87$ ($p < 0.001)$, respectively.

%We show, however, that the correlation is similar to the correlation using all reference native speakers (i.e. $r = -0.87$, $p < 0.001$), hence is not significantly different. Acoustic distances computed per gender (i.e.~using only male or female native and non-native speakers) were also not significantly different from using all speakers, as the correlation obtained using \texttt{w2v2-en}-based neural representations for male and female speakers is $r = -0.86$ $(p < 0.001)$ and $r = -0.87$ $(p < 0.001)$, respectively.

Finally, we calculated the correlation with human perception using \texttt{w2v2-en} when instead of the full 69-word paragraph individual words were selected. These correlations ranged from $r = -0.50$ for the word ``She'' to $r = -0.78$ for the word ``Stella''. The average correlation was $r = -0.67$ ($p < 0.001$, $\sigma = 0.06$). While the results on the basis of the full dataset show a higher correlation with human perception, it is noteworthy that some individual words also appear to correlate strongly with perception.

%several words by themselves show a It is striking that only selecting several 
%While the results are considerably lower compared to using all words in the acoustic distance calculation, these results indicate that the words \textit{Stella}, \textit{her}, \textit{the}, \textit{train}, and \textit{of} strongly correlate with human perception, while the words \textit{need}, \textit{we}, \textit{big}, \textit{five}, and \textit{she} indicate more moderate correlations, presenting information which cannot be clearly seen by using averaged acoustic distances.

% \begin{table}[t]
% \centering
%     \begin{tabular}{lr}
%         \toprule
%         \textbf{Word} & \textbf{Mean \textit{r}}\\
%         \midrule
%         Stella & -0.77 \\
%         Her & -0.77 \\
%         The & -0.77 \\
%         Train & -0.74 \\
%         Of & -0.74 \\
%         \midrule
%         Need & -0.57 \\
%         We & -0.54 \\
%         Big & -0.54 \\
%         Five & -0.53 \\
%         She & -0.45 \\
%         \bottomrule
%     \end{tabular}
% \caption{Pearson correlation coefficients $r$ between individual word-based \texttt{w2v2-en} distances and SAA human native-likeness ratings. All correlations are significant at the $p < 0.001$ level.}
% \label{tab:indivwords}
% \end{table}

\subsection{Understanding representations}
% performance per non-native language background
% performance for different windows of perceptual distances
% non-native speakers within native-likeness range
% regression models
% living experiment
% visualization tool

\begin{figure}[ht!]
  \begin{center}
    \includegraphics[width=3in]{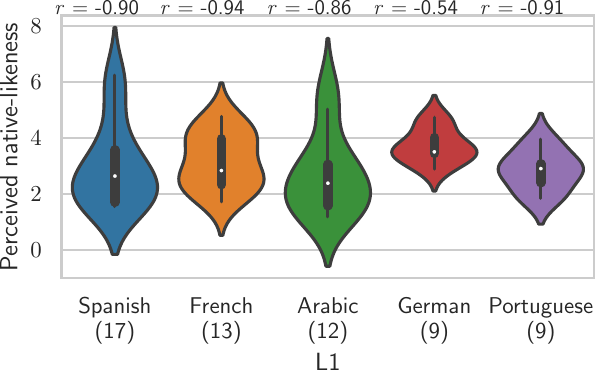}
    \caption{Violin plots visualizing the spread in native-likeness ratings for speakers of different native languages. The number of speakers is indicated between parentheses below the language. The correlation for each group is indicated above each violin plot.}
    \label{fig:L1-ratings}
  \end{center} 
\end{figure}

To assess whether our best-performing model can also identify more fine-grained differences, we evaluate the model against several subsets of data consisting of non-native speakers from the same native language background. 
The spread in native-likeness ratings, as well as the correlations for the groups with the largest number of speakers are shown in Figure~\ref{fig:L1-ratings}.
Except for the native speakers of German (with a relatively restricted range in native-likeness ratings), we observe strong correlations for all groups of speakers.

The low correlation for German speakers suggests that a restricted range of native-likeness ratings may negatively affect the correlation with human perceptual ratings. However, subsequent experiments using \texttt{w2v2-en} (not shown) revealed that the correlation when only including speakers who received average native-likeness ratings between (e.g.,) 5 and 6 was not lower than when increasing the range to include all speakers who received average native-likeness ratings between (e.g.,) 3 and 6.

To identify whether the acoustic distances computed using \texttt{w2v2-en} capture additional pronunciation characteristics compared to acoustic distances based on MFCCs or phonetic transcription-based distances, we fitted a multiple regression model predicting the human native-likeness ratings of the Speech Accent Archive dataset. Table \ref{table:regression} shows the estimated coefficients (for standardized predictors), and summarizes the fit of the regression model. Acoustic distances computed using \texttt{w2v2-en} and phonetic transcription-based distances calculated by the PMI-based Levenshtein distance both contribute significantly to the model fit ($p < 0.05$), whereas this is not the case for the MFCC-based distances. The contribution of \texttt{w2v2-en} is strongest as is clear from the standardized estimates. Overall, this model accounts for 77\% of the variation in the human native-likeness assessments. A model fitted exclusively on the basis of the phonetic transcription-based distances explains 60\% of the variation in the human native-likeness ratings. Given that a model fitted exclusively on the basis of the \texttt{w2v2-en}-based distances explains 76\% of the variation in the human native-likeness ratings, these self-supervised neural models capture information that is not captured by phonetic transcriptions. Nevertheless, the abstractions provided by phonetic transcriptions do provide some (limited) additional information over the self-supervised neural models.

\begin{table*}[ht]
\centering
    \begin{tabular}{lrrrr}
      \toprule
      & \textbf{Estimate (in $z$)} & \textbf{Std. Error} & \textbf{\textit{t}-value} & \textbf{\textit{p}-value}\\
      \midrule
      (Intercept) & 2.98 & 0.03 & 86.56 & $<$ 0.001\\
      LD \citep{wieling2014a} & -0.15 & 0.06 & -2.35 & $<$ 0.05\\
      MFCC \citep{acoustic-measure} & 0.08 & 0.06 & 1.33 & 0.18\\
      \texttt{w2v2-en} & -0.98 & 0.08 & -11.75 & $<$ 0.001\\
      \bottomrule
      \noalign{\vskip 1mm}
    \end{tabular}
  \resizebox{0.9\textwidth}{!}{\centerline{}}
\caption{Coefficients of a multiple regression model ($R^{2} = 0.77$) predicting human native-likeness judgements on the basis of phonetic transcription-based distances computed with the PMI-based Levenshtein distance (LD), and acoustic-only distances based on MFCCs and \texttt{w2v2-en}.}
\label{table:regression}
\end{table*}

Table \ref{table:living} shows how acoustic distances on the basis of the MFCC approach and the \texttt{w2v2-en} model are affected by intonation and timing differences, as well as by recording device. For each condition, ten repetitions were recorded. The recordings are the same as those used by \citet{acoustic-measure}. To enable a better comparison, however, all obtained distances are scaled between 0 and 1. It is clear that the averaged distances from the repetitions of the same word (which may have differed slightly) are somewhat smaller for the \texttt{w2v2-en} model than for the MFCC approach. Importantly, whereas the MFCC approach does not cope well with a different recording device, the \texttt{w2v2-en} model appears to be much more robust (i.e.~resulting in values closer to those for the normal pronunciation). Interestingly, whereas the MFCC approach appears to find larger differences between recordings differing in intonation compared to those with a lengthened first syllable, this is opposite for the \texttt{w2v2-en} model. Both methods, however, appear to be sensitive to differences regarding these aspects. 

\begin{figure*}[ht]
     \centering
      \begin{subfigure}[b]{0.49\textwidth}
         \centering
         \includegraphics[width=3in]{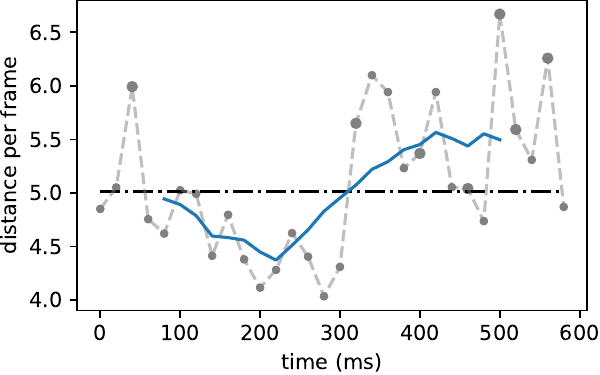}
         \caption{Normal pronunciation}
         \label{fig:vistool-normal}
     \end{subfigure}
     \hfill
     \begin{subfigure}[b]{0.49\textwidth}
         \centering
         \includegraphics[width=3in]{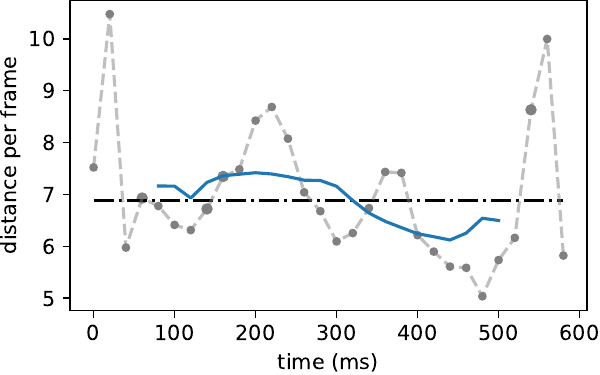}
         \caption{Different recording device}
         \label{fig:fig:vistool-recdev}
     \end{subfigure}
     \par\bigskip
     \begin{subfigure}[b]{0.49\textwidth}
         \centering
         \includegraphics[width=3in]{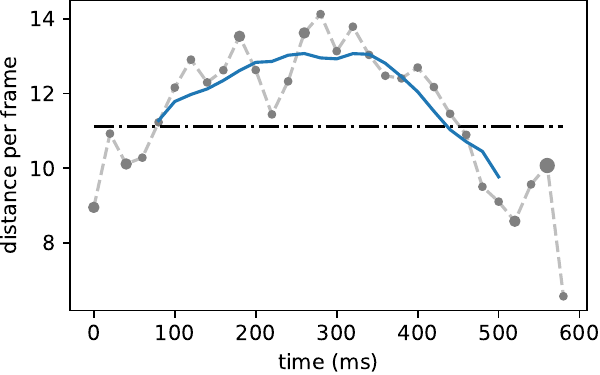}
         \caption{Rising intonation}
         \label{fig:fig:vistool-rising}
      \end{subfigure}
      \begin{subfigure}[b]{0.49\textwidth}
         \centering
         \includegraphics[width=3in]{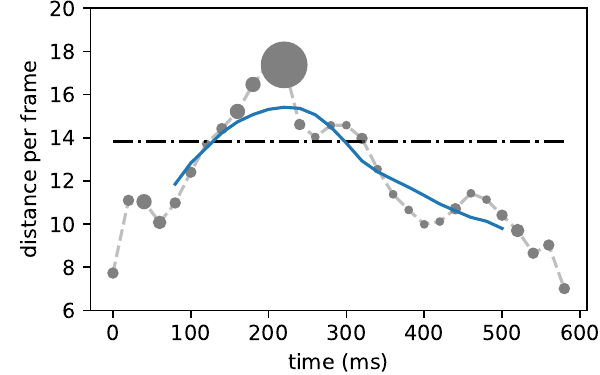}
         \caption{Lengthened first syllable}
         \label{fig:vistool-lengthened}
     \end{subfigure}
     \caption{Visualization of neural acoustic distances per frame (based on \texttt{w2v2-en}) comparing each of the four variants of ``living'' to the same normal pronunciation. The horizontal line represents the global distance value (i.e.~the average of all individual frames). The blue continuous line represents the moving average distance based on 9 frames, corresponding to 180ms. As a result of the moving average, the blue line does not cover the entire duration of the sample. Larger bullet sizes indicate that multiple frames in the reference normal pronunciation are aligned to a single frame in the variant of ``living'' listed on the $x$-axis. Note the different scales of the $y$-axis, reflecting larger differences for the bottom two graphs compared to the top two graphs. See the text for further details.}
     \label{fig:fourplots}
\end{figure*}

% finally, show that \texttt{w2v2-en} representations computed using the best-performing output layer capture both intonation and timing differences, similar to using MFCCs. The \texttt{w2v2-en}-based acoustic distances increase when we compare pronunciations with rising intonation or a lengthened first syllable to standard pronunciations of 'living'. Particularly remarkable are the acoustic differences computed between the standard pronunciation recorded with two different microphones, showing a substantially lower increase for \texttt{w2v2-en}-based distances compared to MFCC-based distances.

\begin{table*}[ht]
\centering
  \begin{tabular}{lrr}
    \toprule
     & \textbf{\texttt{w2v2-en}} & \textbf{MFCC} \\
    \midrule
    Normal pronunciation & 0.18 (0.10) & 0.23 (0.13) \\
    Normal pronunciation (different recording device) & 0.29 (0.08) & 0.88 (0.04) \\
    Rising intonation & 0.61 (0.07) & 0.92 (0.03) \\
    Lengthened first syllable & 0.91 (0.05) & 0.80 (0.03) \\
    \bottomrule
  \end{tabular}
\caption{Normalized averaged acoustic distances of four variants of the word ``living'' (each repeated ten times) compared to the normal pronunciation of ``living'', computed using \texttt{w2v2-en} and MFCCs. Standard deviations are shown between parentheses.}
\label{table:living}
\end{table*}

For illustration, Figure~\ref{fig:fourplots} visualizes a comparison between a single normal pronunciation of ``living'' and four other pronunciations. Specifically, Figure~\ref{fig:vistool-normal} shows a comparison with another normal pronunciation. Figure~\ref{fig:fig:vistool-recdev} shows a comparison with the same pronunciation, but using a different recording device. Figure~\ref{fig:fig:vistool-rising} shows a comparison with a rising intonation pronunciation. Finally, Figure~\ref{fig:vistool-lengthened} shows a comparison with a lengthened first syllable pronunciation. In line with Table~\ref{table:living}, the values on the $y$-axis show that the distance between the two normal pronunciations is smaller than when using a different recording device. Note that these distances were not normalized, as they simply compare two recordings. Both distances, however, are smaller than comparing against rising intonation (revealing a curvilinear pattern) and a lengthened first syllable (showing the largest difference at the beginning of the word; the lengthening is clear from the larger circle denoting an alignment with similar samples differing in duration).

\section{Discussion and conclusion}

In this study, we investigated how several self-supervised neural models may be used to automatically quantify pronunciation variation without needing to use phonetic transcription-based approaches.
We used neural representations to calculate word-based pronunciation differences for English accents and Norwegian dialects, and compared the results to human perceptual judgements. While these ratings were provided on relatively crude (5 to 10-point) scales, and individual raters' biases or strategies may have affected their ratings, averaging across a large number of raters for each sample likely yields an adequate estimate of native-likeness or perceived dialect distance. Our experiments showed that acoustic distances computed with Transformer-based models, such as \texttt{w2v2-en}, closely match the averaged human native-likeness ratings for the English datasets, and that performance greatly depended on the choice of layer. This finding not only demonstrates that these layers contain useful abstractions and generalizations of acoustic information, but also shows that the final layers represent information that is tailored to the target objective (which was speech recognition instead of our present goal of quantifying acoustic differences). This result is in line with findings in the field of natural language processing when using Transformer-based methods with textual data \citep{tenneyBERTRediscoversClassical2019, devriesWhatSpecialBERT2020}. Furthermore, the \texttt{w2v2} and \texttt{XLSR} models appeared to be robust against the choice of reference speaker(s) to compare against. Even choosing a single reference speaker resulted in correlations that were not substantially different from those that used the full set. Interestingly, correlations on the basis of some words were not much lower than those on the basis of the full set of words, suggesting that a smaller number of words may already yield an adequate assessment of native-likeness. 

Our newly-developed visualization tool helped us to understand these `black box' models, as the visualization showed where the differences between two pronunciations were largest (i.e.~the locus of the effect). This type of tool could potentially be used to provide visual feedback to learners of a second language or people with a speech disorder. However, the actual effectiveness of such an approach would need to be investigated.

Our results seem to indicate that phonetic transcriptions are no longer essential when the goal is to use these to quantify how different non-native speech is from native speech, and an appropriate Transformer-based model is available. This suggests that a time-consuming and labor intensive process can be omitted in this case. While our regression model showed that phonetic transcriptions did offer additional information not present in our neural acoustic-only approach, this information gain was very limited (an increase in $R^2$ of only one percent). We furthermore showed that our neural method captures aspects of pronunciations (such as subtle durational or intonation differences) that are hard to capture by a set of discrete symbols used in phonetic transcriptions. Importantly, in contrast to a previous relatively successful acoustic approach \citep{acoustic-measure}, our present neural acoustic approach is relatively unaffected by non-linguistic variation (i.e.~caused by using a different recording device). Nevertheless, further detailed research is needed to obtain a better view of what phonetic information is (not) captured by these models.

% Additionally, human raters might as well be no longer necessary for assessing the reliability and accuracy of our acoustic-only pronunciation distance measure, as we found that the perceptual ratings were reliable, and acoustic distances strongly match the perceptual distances from different datasets.
% Additionally, the qualitative comparison of the visualizations based on Dutch dialect pronunciations revealed that traditional dialect differences are visible when neural representations from a neural model fine-tuned on English are used, showing additional proof that validation against human perceptual ratings could be optional.

% Moreover, we demonstrated that Transformer-based neural speech representations not only capture segmental differences, but, in contrast to phonetic transcriptions, also automatically include intonational and durational information, while being minimally affected by non-linguistic variation caused by using a different recording device.

In contrast to the performance on the English datasets, we found that Transformer-based neural representations performed worse when applied to the Norwegian dialects dataset. However, pronunciations of the Norwegian dialects dataset were represented by a model which was trained exclusively or dominantly on English speech. Unfortunately, Norwegian was not among the pre-training languages included in the multilingual (\texttt{XLSR}) model, nor available for fine-tuning. We expect to see improved performance for a Norwegian \texttt{w2v2} model (when made available). Unfortunately, creating such a model is very costly in terms of required resources (generally based on hundreds of hours of speech) and computing power.
At present, we estimate that pre-training (even without hyperparameter tuning and optimization) a new \texttt{w2v2} model for a different language takes about 150 days on a single state-of-the-art NVIDIA A100 GPU (costing approximately US\$ 10,000). Using multiple GPUs in parallel reduces this duration, but also increases the required costs. Fortunately, the cost of these GPUs will usually decrease over time, and the speed of newly developed GPUs will increase.

% While we were not able to evaluate the performance of the Dutch dialects dataset against perceptual ratings, the quality of these recordings was rather low and would likely not have resulted in a high performance, as even the \texttt{w2v2-en} model appears to be affected by non-speech related factors (such as the microphone used). The relatively low explained variance of the MDS plot on the basis of the \texttt{w2v2-en} model as compared to using the PMI-based Levenshtein distance algorithm also supports this. Nevertheless, also for the \texttt{w2v2-en} model, the visualization did appear to somewhat distinguish the major Dutch dialect areas. Interestingly, the \texttt{XLSR-nl} model, which included Dutch both during the pre-training and during the fine-tuning stage, did not appear to improve over the monolingual English \texttt{w2v2-en} model. However, we cannot exclude the possibility that the sub-optimal audio quality may have limited the ability of the neural models to identify sufficient signal in the noise. 

Even though the evaluated architectures were originally designed for natural language processing, and the specific acoustic self-supervised neural models were created for improving performance in the domain of transforming speech to text, we have shown that the neural representations can also be successfully applied to an unrelated task in a different domain.
Moreover, we have illustrated that Transformer-based speech representations are able to model fine-grained differences in homogeneous speaker groups (e.g., from the same language background), and adequately generalize over individual speaker differences, including gender, which makes them potentially useful for other tasks as well.

While our results are promising, the application of the \texttt{w2v2} approach for modeling pronunciation differences is only possible when an existing \texttt{w2v2} model is available for the language in question, or when sufficient data and computing resources are available to create a new \texttt{w2v2} model. In contrast to creating a new \texttt{w2v2} model for a new language, adjusting an existing model for a different task (such as quantifying differences with respect to pitch contours or timing patterns) is easier. This only requires an existing model to be fine-tuned on labeled examples. Generally, the amount of labeled data (and required GPU time) needed for fine-tuning is considerably lower compared to the large quantities of (unlabeled) data needed for pre-training. However, if such resources are not available, the monolingual English model appears to be a suitable alternative, generally outperforming the language-invariant acoustic-only method proposed by \citet{acoustic-measure}. Future work, however, should be aimed at further investigating how existing high-resource language \texttt{w2v2} models may be exploited or extended when analyzing language variation in low-resource languages.

\section*{Declaration of interest}
Declarations of interest: none.

\section*{Acknowledgments}
The authors thank Hedwig Sekeres for creating the transcriptions of the Dutch speakers dataset, and Anna Pot for creating the visualization of the acoustic distance measure.  We furthermore thank three anonymous reviewers and the editor for their insightful comments, which have allowed us to improve this paper.

\newpage
\bibliographystyle{acl_natbib}
\bibliography{references}

\begin{thebibliography}{81}
\expandafter\ifx\csname natexlab\endcsname\relax\def\natexlab#1{#1}\fi

\bibitem[{Ardila et~al.(2020)Ardila, Branson, Davis, Kohler, Meyer, Henretty,
  Morais, Saunders, Tyers, and Weber}]{ardila2019common}
Rosana Ardila, Megan Branson, Kelly Davis, Michael Kohler, Josh Meyer, Michael
  Henretty, Reuben Morais, Lindsay Saunders, Francis Tyers, and Gregor Weber.
  2020.
\newblock \href {https://aclanthology.org/2020.lrec-1.520} {Common voice: A
  massively-multilingual speech corpus}.
\newblock In \emph{Proceedings of the 12th Language Resources and Evaluation
  Conference}, pages 4218--4222, Marseille, France. European Language Resources
  Association.

\bibitem[{Asher and García(1969)}]{asher1969optimal}
James~J. Asher and Ramiro García. 1969.
\newblock \href {http://www.jstor.org/stable/323026} {The optimal age to learn
  a foreign language}.
\newblock \emph{The Modern Language Journal}, 53(5):334--341.

\bibitem[{Babu et~al.(2021)Babu, Wang, Tjandra, Lakhotia, Xu, Goyal, Singh, von
  Platen, Saraf, Pino, Baevski, Conneau, and Auli}]{babu2021xlsr}
Arun Babu, Changhan Wang, Andros Tjandra, Kushal Lakhotia, Qiantong Xu, Naman
  Goyal, Kritika Singh, Patrick von Platen, Yatharth Saraf, Juan Pino, Alexei
  Baevski, Alexis Conneau, and Michael Auli. 2021.
\newblock \href {http://arxiv.org/abs/2111.09296} {{XLS-R: Self-supervised
  Cross-lingual Speech Representation Learning at Scale}}.

\bibitem[{Baevski et~al.(2020{\natexlab{a}})Baevski, Schneider, and
  Auli}]{baevski2019vq}
Alexei Baevski, Steffen Schneider, and Michael Auli. 2020{\natexlab{a}}.
\newblock \href {http://arxiv.org/abs/1910.05453} {vq-wav2vec: Self-supervised
  learning of discrete speech representations}.

\bibitem[{Baevski et~al.(2020{\natexlab{b}})Baevski, Zhou, Mohamed, and
  Auli}]{baevski2020wav2vec}
Alexei Baevski, Yuhao Zhou, Abdelrahman Mohamed, and Michael Auli.
  2020{\natexlab{b}}.
\newblock \href
  {https://proceedings.neurips.cc/paper/2020/file/92d1e1eb1cd6f9fba3227870bb6d7f07-Paper.pdf}
  {wav2vec 2.0: A framework for self-supervised learning of speech
  representations}.
\newblock In \emph{Advances in Neural Information Processing Systems},
  volume~33, pages 12449--12460. Curran Associates, Inc.

\bibitem[{Bartelds et~al.(2020)Bartelds, Richter, Liberman, and
  Wieling}]{acoustic-measure}
Martijn Bartelds, Caitlin Richter, Mark Liberman, and Martijn Wieling. 2020.
\newblock \href {https://doi.org/10.3389/frai.2020.00039} {A new acoustic-based
  pronunciation distance measure}.
\newblock \emph{Frontiers in Artificial Intelligence}, 3:39.

\bibitem[{Blodgett et~al.(2016)Blodgett, Green, and
  O{'}Connor}]{blodgett2016demographic}
Su~Lin Blodgett, Lisa Green, and Brendan O{'}Connor. 2016.
\newblock \href {https://doi.org/10.18653/v1/D16-1120} {Demographic dialectal
  variation in social media: A case study of {A}frican-{A}merican {E}nglish}.
\newblock In \emph{Proceedings of the 2016 Conference on Empirical Methods in
  Natural Language Processing}, pages 1119--1130, Austin, Texas. Association
  for Computational Linguistics.

\bibitem[{Boberg(2010)}]{boberg2010english}
Charles Boberg. 2010.
\newblock \href {https://doi.org/10.1017/CBO9780511781056} {\emph{The English
  Language in Canada: Status, History and Comparative Analysis}}.
\newblock Studies in English Language. Cambridge University Press.

\bibitem[{Bucholtz(2007)}]{bucholtz2007variation}
Mary Bucholtz. 2007.
\newblock \href {http://www.jstor.org/stable/24049459} {Variation in
  transcription}.
\newblock \emph{Discourse Studies}, 9(6):784--808.

\bibitem[{Chen et~al.(2016)Chen, Wee, Tong, Ma, and Li}]{chen2016large}
Nancy~F. Chen, Darren Wee, Rong Tong, Bin Ma, and Haizhou Li. 2016.
\newblock \href {https://doi.org/https://doi.org/10.1016/j.specom.2016.07.005}
  {Large-scale characterization of non-native mandarin chinese spoken by
  speakers of european origin: Analysis on icall}.
\newblock \emph{Speech Communication}, 84:46--56.

\bibitem[{Church and Hanks(1990)}]{church1990word}
Kenneth~Ward Church and Patrick Hanks. 1990.
\newblock \href {https://doi.org/10.5555/89086.89095} {Word association norms,
  mutual information, and lexicography}.
\newblock \emph{Computational Linguistics}, 16(1):22–29.

\bibitem[{Conneau et~al.(2020)Conneau, Baevski, Collobert, Mohamed, and
  Auli}]{conneau2020unsupervised}
Alexis Conneau, Alexei Baevski, Ronan Collobert, Abdelrahman Mohamed, and
  Michael Auli. 2020.
\newblock \href {http://arxiv.org/abs/2006.13979} {Unsupervised cross-lingual
  representation learning for speech recognition}.

\bibitem[{Cronbach(1951)}]{cronbach1951coefficient}
Lee~J. Cronbach. 1951.
\newblock \href {https://doi.org/10.1007/BF02310555} {Coefficient alpha and the
  internal structure of tests}.
\newblock \emph{Psychometrika}, 16(3):297--334.

\bibitem[{Cucchiarini(1996)}]{cucchiarini1996assessing}
Catia Cucchiarini. 1996.
\newblock \href {https://doi.org/10.3109/02699209608985167} {Assessing
  transcription agreement: Methodological aspects}.
\newblock \emph{Clinical Linguistics \& Phonetics}, 10(2):131--155.

\bibitem[{Devlin et~al.(2019)Devlin, Chang, Lee, and
  Toutanova}]{devlin_bert_2019}
Jacob Devlin, Ming-Wei Chang, Kenton Lee, and Kristina Toutanova. 2019.
\newblock \href {https://doi.org/10.18653/v1/N19-1423} {{{BERT}}:
  {{Pre}}-training of {{Deep Bidirectional Transformers}} for {{Language
  Understanding}}}.
\newblock In \emph{Proceedings of the 2019 {{Conference}} of the {{North
  American Chapter}} of the {{Association}} for {{Computational Linguistics}}:
  {{Human Language Technologies}}, {{Volume}} 1 ({{Long}} and {{Short
  Papers}})}, pages 4171--4186, {Minneapolis, Minnesota}. {Association for
  Computational Linguistics}.

\bibitem[{Duckworth et~al.(1990)Duckworth, Allen, Hardcastle, and
  Ball}]{duckworth1990extensions}
Martin Duckworth, George Allen, William Hardcastle, and Martin Ball. 1990.
\newblock \href {https://doi.org/10.3109/02699209008985489} {Extensions to the
  international phonetic alphabet for the transcription of atypical speech}.
\newblock \emph{Clinical Linguistics \& Phonetics}, 4(4):273--280.

\bibitem[{Ferragne and Pellegrino(2010)}]{Ferragne2010}
Emmanuel Ferragne and François Pellegrino. 2010.
\newblock \href {https://doi.org/https://doi.org/10.1016/j.wocn.2010.07.002}
  {Vowel systems and accent similarity in the british isles: Exploiting
  multidimensional acoustic distances in phonetics}.
\newblock \emph{Journal of Phonetics}, 38(4):526--539.

\bibitem[{Flege(1988)}]{flege1988factors}
James~Emil Flege. 1988.
\newblock \href {https://doi.org/10.1121/1.396876} {Factors affecting degree of
  perceived foreign accent in english sentences}.
\newblock \emph{The Journal of the Acoustical Society of America},
  84(1):70—79.

\bibitem[{Gales et~al.(2014)Gales, Knill, Ragni, and Rath}]{gales2014speech}
Mark~JF Gales, Kate~M Knill, Anton Ragni, and Shakti~P Rath. 2014.
\newblock \href {https://eprints.whiterose.ac.uk/152840/} {Speech recognition
  and keyword spotting for low-resource languages: Babel project research at
  cued}.
\newblock In \emph{Fourth International workshop on spoken language
  technologies for under-resourced languages (SLTU-2014)}, pages 16--23.
  International Speech Communication Association (ISCA).

\bibitem[{Garofalo et~al.(2007)Garofalo, Graff, Paul, and
  Pallett}]{garofalo2007csr}
John~S. Garofalo, David Graff, Dough Paul, and David Pallett. 2007.
\newblock \href {https://doi.org/https://doi.org/10.35111/ewkm-cg47} {{CSR-I
  (WSJ0) Complete LDC93S6A}}.
\newblock \emph{Web Download. Philadelphia: Linguistic Data Consortium}.

\bibitem[{Giorgino(2009)}]{giorgino2009}
Toni Giorgino. 2009.
\newblock \href {https://doi.org/10.18637/jss.v031.i07} {{Computing and
  Visualizing Dynamic Time Warping Alignments in R: The dtw Package}}.
\newblock \emph{Journal of Statistical Software}, 31(7):1–24.

\bibitem[{Gooskens(2005)}]{gooskens2005}
Charlotte Gooskens. 2005.
\newblock \href {https://doi.org/10.1017/S0332586505001319} {{How well can
  Norwegians identify their dialects?}}
\newblock \emph{Nordic Journal of Linguistics}, 28(1):37–60.

\bibitem[{Gooskens and Heeringa(2004)}]{gooskens2004perceptive}
Charlotte Gooskens and Wilbert Heeringa. 2004.
\newblock \href {https://doi.org/10.1017/S0954394504163023} {Perceptive
  evaluation of levenshtein dialect distance measurements using norwegian
  dialect data}.
\newblock \emph{Language Variation and Change}, 16(3):189–207.

\bibitem[{Graves et~al.(2006)Graves, Fern\'{a}ndez, Gomez, and
  Schmidhuber}]{graves2006connectionist}
Alex Graves, Santiago Fern\'{a}ndez, Faustino Gomez, and J\"{u}rgen
  Schmidhuber. 2006.
\newblock \href {https://doi.org/10.1145/1143844.1143891} {Connectionist
  temporal classification: Labelling unsegmented sequence data with recurrent
  neural networks}.
\newblock In \emph{Proceedings of the 23rd International Conference on Machine
  Learning}, ICML '06, page 369–376, New York, NY, USA. Association for
  Computing Machinery.

\bibitem[{Hakkani-T{\"u}r et~al.(2002)Hakkani-T{\"u}r, Riccardi, and
  Gorin}]{hakkani2002active}
Dilek Hakkani-T{\"u}r, Giuseppe Riccardi, and Allen Gorin. 2002.
\newblock \href {https://doi.org/10.1109/ICASSP.2002.5745510} {Active learning
  for automatic speech recognition}.
\newblock In \emph{2002 IEEE International Conference on Acoustics, Speech, and
  Signal Processing}, volume~4, pages IV--3904. IEEE.

\bibitem[{Hanani et~al.(2013)Hanani, Russell, and Carey}]{hanani2013human}
Abualsoud Hanani, Martin~J Russell, and Michael~J Carey. 2013.
\newblock \href {https://doi.org/https://doi.org/10.1016/j.csl.2012.01.003}
  {Human and computer recognition of regional accents and ethnic groups from
  british english speech}.
\newblock \emph{Computer Speech \& Language}, 27(1):59--74.
\newblock Special issue on Paralinguistics in Naturalistic Speech and Language.

\bibitem[{Heeringa et~al.(2009)Heeringa, Johnson, and
  Gooskens}]{heeringa2009measuring}
Wilbert Heeringa, Keith Johnson, and Charlotte Gooskens. 2009.
\newblock \href {https://doi.org/https://doi.org/10.1016/j.specom.2008.07.006}
  {Measuring norwegian dialect distances using acoustic features}.
\newblock \emph{Speech Communication}, 51(2):167--183.

\bibitem[{Heeringa(2004)}]{heeringa2004measuring}
{Wilbert Jan} Heeringa. 2004.
\newblock \href
  {https://research.rug.nl/en/publications/measuring-dialect-pronunciation-differences-using-levenshtein-dis}
  {\emph{Measuring Dialect Pronunciation Differences using Levenshtein
  Distance}}.
\newblock Ph.D. thesis, University of Groningen.

\bibitem[{Huang et~al.(2014{\natexlab{a}})Huang, Slaney, Seltzer, and
  Gong}]{huang2014towards}
Yan Huang, Malcolm Slaney, Michael~L Seltzer, and Yifan Gong.
  2014{\natexlab{a}}.
\newblock \href
  {https://www.isca-speech.org/archive/interspeech_2014/i14_0845.html} {Towards
  better performance with heterogeneous training data in acoustic modeling
  using deep neural networks}.
\newblock In \emph{Fifteenth Annual Conference of the International Speech
  Communication Association}.

\bibitem[{Huang et~al.(2014{\natexlab{b}})Huang, Yu, Liu, and
  Gong}]{huang2014comparative}
Yan Huang, Dong Yu, Chaojun Liu, and Yifan Gong. 2014{\natexlab{b}}.
\newblock \href
  {https://www.isca-speech.org/archive/interspeech_2014/i14_1895.html} {A
  comparative analytic study on the gaussian mixture and context dependent deep
  neural network hidden markov models}.
\newblock In \emph{Fifteenth Annual Conference of the International Speech
  Communication Association}.

\bibitem[{Huckvale(2007)}]{Huckvale2007}
Mark Huckvale. 2007.
\newblock \href {https://doi.org/10.1007/978-3-540-74122-0_20} {\emph{ACCDIST:
  An Accent Similarity Metric for Accent Recognition and Diagnosis}}, pages
  258--275. Springer Berlin Heidelberg, Berlin, Heidelberg.

\bibitem[{Jang et~al.(2017)Jang, Gu, and Poole}]{jang_categorical_2017}
Eric Jang, Shixiang Gu, and Ben Poole. 2017.
\newblock \href {http://arxiv.org/abs/1611.01144} {Categorical
  {{Reparameterization}} with {{Gumbel}}-{{Softmax}}}.
\newblock \emph{arXiv:1611.01144 [cs, stat]}.

\bibitem[{Jeszenszky et~al.(2017)Jeszenszky, Stoeckle, Glaser, and
  Weibel}]{jeszenszky2017exploring}
P{\'e}ter Jeszenszky, Philipp Stoeckle, Elvira Glaser, and Robert Weibel. 2017.
\newblock \href {https://doi.org/10.1017/jlg.2017.5} {Exploring global and
  local patterns in the correlation of geographic distances and morphosyntactic
  variation in swiss german}.
\newblock \emph{Journal of Linguistic Geography}, 5(2):86–108.

\bibitem[{Kahn et~al.(2020)Kahn, Rivi{\`e}re, Zheng, Kharitonov, Xu,
  Mazar{\'e}, Karadayi, Liptchinsky, Collobert, Fuegen et~al.}]{kahn2020libri}
Jacob Kahn, Morgane Rivi{\`e}re, Weiyi Zheng, Evgeny Kharitonov, Qiantong Xu,
  Pierre-Emmanuel Mazar{\'e}, Julien Karadayi, Vitaliy Liptchinsky, Ronan
  Collobert, Christian Fuegen, et~al. 2020.
\newblock \href {https://doi.org/10.1109/ICASSP40776.2020.9052942}
  {{Libri-Light: A Benchmark for ASR with Limited or No Supervision}}.
\newblock In \emph{ICASSP 2020-2020 IEEE International Conference on Acoustics,
  Speech and Signal Processing (ICASSP)}, pages 7669--7673. IEEE.

\bibitem[{Koenecke et~al.(2020)Koenecke, Nam, Lake, Nudell, Quartey, Mengesha,
  Toups, Rickford, Jurafsky, and Goel}]{koenecke2020}
Allison Koenecke, Andrew Nam, Emily Lake, Joe Nudell, Minnie Quartey, Zion
  Mengesha, Connor Toups, John~R. Rickford, Dan Jurafsky, and Sharad Goel.
  2020.
\newblock \href {https://doi.org/10.1073/pnas.1915768117} {{Racial disparities
  in automated speech recognition}}.
\newblock \emph{Proceedings of the National Academy of Sciences},
  117(14):7684--7689.

\bibitem[{Koster and Koet(1993)}]{koster1993evaluation}
Cor~J Koster and Ton Koet. 1993.
\newblock \href {https://doi.org/10.1111/j.1467-1770.1993.tb00173.x} {{The
  evaluation of accent in the English of Dutchmen}}.
\newblock \emph{Language learning}, 43(1):69--92.

\bibitem[{Leather(1983)}]{leather1983second}
Jonathan Leather. 1983.
\newblock \href {https://doi.org/10.1017/S0261444800010120} {Second-language
  pronunciation learning and teaching}.
\newblock \emph{Language Teaching}, 16(3):198--219.

\bibitem[{Levenshtein(1966)}]{levenshtein}
Vladimir Levenshtein. 1966.
\newblock Binary codes capable of correcting deletions, insertions, and
  reversals.
\newblock \emph{Cybernetics and Control Theory}, 10(8):707--710.

\bibitem[{Liberman(2018)}]{liberman2018}
Mark Liberman. 2018.
\newblock \href {https://doi.org/10.7208/chicago/9780226562599.003.0009}
  {Towards progress in theories of language sound structure}.
\newblock In Diane Brentari and Jackson~L. Lee, editors, \emph{Shaping
  phonology}. University of Chicago Press.

\bibitem[{Ling et~al.(2020)Ling, Liu, Salazar, and Kirchhoff}]{ling2020deep}
Shaoshi Ling, Yuzong Liu, Julian Salazar, and Katrin Kirchhoff. 2020.
\newblock \href {https://doi.org/10.1109/ICASSP40776.2020.9053176} {Deep
  contextualized acoustic representations for semi-supervised speech
  recognition}.
\newblock In \emph{ICASSP 2020 - 2020 IEEE International Conference on
  Acoustics, Speech and Signal Processing (ICASSP)}, pages 6429--6433.

\bibitem[{Livescu and Glass(2000)}]{livescu2000lexical}
K.~Livescu and J.~Glass. 2000.
\newblock \href {https://doi.org/10.1109/ICASSP.2000.862074} {Lexical modeling
  of non-native speech for automatic speech recognition}.
\newblock In \emph{2000 IEEE International Conference on Acoustics, Speech, and
  Signal Processing. Proceedings (Cat. No.00CH37100)}, volume~3, pages
  1683--1686.

\bibitem[{Ma et~al.(2021)Ma, Ryant, and Liberman}]{ma:21}
Danni Ma, Neville Ryant, and Mark Liberman. 2021.
\newblock \href {https://doi.org/10.1109/ICASSP39728.2021.9414776} {Probing
  acoustic representations for phonetic properties}.
\newblock In \emph{ICASSP 2021 - 2021 IEEE International Conference on
  Acoustics, Speech and Signal Processing (ICASSP)}, pages 311--315.

\bibitem[{Magen(1998)}]{magen1998perception}
Harriet~S. Magen. 1998.
\newblock \href {https://doi.org/https://doi.org/10.1006/jpho.1998.0081} {The
  perception of foreign-accented speech}.
\newblock \emph{Journal of Phonetics}, 26(4):381--400.

\bibitem[{Mathur et~al.(2019)Mathur, Isopoussu, Kawsar, Berthouze, and
  Lane}]{mathur2019}
Akhil Mathur, Anton Isopoussu, Fahim Kawsar, Nadia Berthouze, and Nicholas~D.
  Lane. 2019.
\newblock \href {https://doi.org/10.1145/3302506.3310398} {{Mic2Mic: Using
  Cycle-Consistent Generative Adversarial Networks to Overcome Microphone
  Variability in Speech Systems}}.
\newblock In \emph{Proceedings of the 18th International Conference on
  Information Processing in Sensor Networks}, IPSN '19, page 169–180, New
  York, NY, USA. Association for Computing Machinery.

\bibitem[{Mermelstein(1976)}]{mermelstein1976distance}
Paul Mermelstein. 1976.
\newblock Distance measures for speech recognition, psychological and
  instrumental.
\newblock \emph{Pattern recognition and artificial intelligence}, 116:374--388.

\bibitem[{Mikolov et~al.(2013)Mikolov, Sutskever, Chen, Corrado, and
  Dean}]{mikolov_distributed_2013}
Tomas Mikolov, Ilya Sutskever, Kai Chen, Greg~S Corrado, and Jeff Dean. 2013.
\newblock \href
  {https://proceedings.neurips.cc/paper/2013/file/9aa42b31882ec039965f3c4923ce901b-Paper.pdf}
  {Distributed representations of words and phrases and their
  compositionality}.
\newblock In \emph{Advances in Neural Information Processing Systems},
  volume~26. Curran Associates, Inc.

\bibitem[{Munro(1995)}]{munro1995nonsegmental}
Murray~J Munro. 1995.
\newblock \href {https://doi.org/10.1017/S0272263100013735} {Nonsegmental
  factors in foreign accent: Ratings of filtered speech}.
\newblock \emph{Studies in Second Language Acquisition}, 17(1):17--34.

\bibitem[{Munro and Derwing(1994)}]{munro1994evaluations}
Murray~J Munro and Tracey~M Derwing. 1994.
\newblock \href {https://doi.org/10.1177/026553229401100302} {Evaluations of
  foreign accent in extemporaneous and read material}.
\newblock \emph{Language Testing}, 11(3):253--266.

\bibitem[{Munro and Derwing(2001)}]{munro2001modeling}
Murray~J Munro and Tracey~M Derwing. 2001.
\newblock \href {https://doi.org/10.1017/S0272263101004016} {Modeling
  perceptions of the accentedness and comprehensibility of l2 speech the role
  of speaking rate}.
\newblock \emph{Studies in second language acquisition}, 23(4):451--468.

\bibitem[{Najafian et~al.(2014)Najafian, DeMarco, Cox, and
  Russell}]{najafian2014unsupervised}
Maryam Najafian, Andrea DeMarco, Stephen Cox, and Martin Russell. 2014.
\newblock \href
  {https://www.isca-speech.org/archive/interspeech_2014/i14_2967.html}
  {Unsupervised model selection for recognition of regional accented speech}.
\newblock In \emph{Fifteenth annual conference of the international speech
  communication association}.

\bibitem[{Nerbonne and Heeringa(1997)}]{nerbonne1997measuring}
John Nerbonne and Wilbert Heeringa. 1997.
\newblock \href {https://aclanthology.org/W97-1102} {Measuring dialect distance
  phonetically}.
\newblock In \emph{Computational Phonology: Third Meeting of the {ACL} Special
  Interest Group in Computational Phonology}.

\bibitem[{Novotney and Callison-Burch(2010)}]{novotney2010cheap}
Scott Novotney and Chris Callison-Burch. 2010.
\newblock \href {https://aclanthology.org/N10-1024} {Cheap, fast and good
  enough: Automatic speech recognition with non-expert transcription}.
\newblock In \emph{Human Language Technologies: The 2010 Annual Conference of
  the North {A}merican Chapter of the Association for Computational
  Linguistics}, pages 207--215, Los Angeles, California. Association for
  Computational Linguistics.

\bibitem[{Nunnally(1978)}]{nunnally1978}
Jum~C. Nunnally. 1978.
\newblock \emph{Psychometric theory}, 2nd ed. edition.
\newblock {McGraw}-{Hill} series in psychology. McGraw-Hill, New York ;.

\bibitem[{Offrede et~al.(2020)Offrede, Jacobi, Rebernik, de~Jong, Keulen,
  Veenstra, Noiray, and Wieling}]{wielinglowlands}
Tom~F. Offrede, Jidde Jacobi, Teja Rebernik, Lisanne de~Jong, Stefanie Keulen,
  Pauline Veenstra, Aude Noiray, and Martijn Wieling. 2020.
\newblock \href {https://doi.org/10.1177/0023830920953169} {The impact of
  alcohol on l1 versus l2}.
\newblock \emph{Language and Speech}.
\newblock PMID: 32856992.

\bibitem[{van~den Oord et~al.(2019)van~den Oord, Li, and
  Vinyals}]{oord2018representation}
Aaron van~den Oord, Yazhe Li, and Oriol Vinyals. 2019.
\newblock \href {http://arxiv.org/abs/1807.03748} {Representation learning with
  contrastive predictive coding}.

\bibitem[{van~den Oord et~al.(2017)van~den Oord, Vinyals, and
  kavukcuoglu}]{van_den_oord_neural_2017}
Aaron van~den Oord, Oriol Vinyals, and koray kavukcuoglu. 2017.
\newblock \href
  {https://proceedings.neurips.cc/paper/2017/file/7a98af17e63a0ac09ce2e96d03992fbc-Paper.pdf}
  {Neural discrete representation learning}.
\newblock In \emph{Advances in Neural Information Processing Systems},
  volume~30. Curran Associates, Inc.

\bibitem[{Panayotov et~al.(2015)Panayotov, Chen, Povey, and
  Khudanpur}]{panayotov_librispeech_2015}
Vassil Panayotov, Guoguo Chen, Daniel Povey, and Sanjeev Khudanpur. 2015.
\newblock \href {https://doi.org/10.1109/ICASSP.2015.7178964} {Librispeech:
  {{An ASR}} corpus based on public domain audio books}.
\newblock In \emph{2015 {{IEEE International Conference}} on {{Acoustics}},
  {{Speech}} and {{Signal Processing}} ({{ICASSP}})}, pages 5206--5210.

\bibitem[{Paul and Baker(1992)}]{paul-baker-1992-design}
Douglas~B. Paul and Janet~M. Baker. 1992.
\newblock \href {https://www.aclweb.org/anthology/H92-1073} {The design for the
  {W}all {S}treet {J}ournal-based {CSR} corpus}.
\newblock In \emph{Speech and Natural Language: Proceedings of a Workshop Held
  at Harriman, New York, {F}ebruary 23-26, 1992}.

\bibitem[{Peters et~al.(2018)Peters, Neumann, Iyyer, Gardner, Clark, Lee, and
  Zettlemoyer}]{petersDeepContextualizedWord2018}
Matthew Peters, Mark Neumann, Mohit Iyyer, Matt Gardner, Christopher Clark,
  Kenton Lee, and Luke Zettlemoyer. 2018.
\newblock \href {https://doi.org/10.18653/v1/N18-1202} {Deep {{Contextualized
  Word Representations}}}.
\newblock In \emph{Proceedings of the 2018 {{Conference}} of the {{North
  American Chapter}} of the {{Association}} for {{Computational Linguistics}}:
  {{Human Language Technologies}}, {{Volume}} 1 ({{Long Papers}})}, pages
  2227--2237, {New Orleans, Louisiana}. {Association for Computational
  Linguistics}.

\bibitem[{Pratap et~al.(2020)Pratap, Xu, Sriram, Synnaeve, and
  Collobert}]{pratap2020mls}
Vineel Pratap, Qiantong Xu, Anuroop Sriram, Gabriel Synnaeve, and Ronan
  Collobert. 2020.
\newblock \href {https://doi.org/10.21437/Interspeech.2020-2826} {{MLS: A
  Large-Scale Multilingual Dataset for Speech Research}}.
\newblock In \emph{Proc. Interspeech 2020}, pages 2757--2761.

\bibitem[{Preston(1999)}]{long1999}
Dennis~Richard Preston. 1999.
\newblock \href
  {http://search.ebscohost.com.proxy-ub.rug.nl/login.aspx?direct=true&db=nlebk&AN=253518&site=ehost-live&scope=site}
  {\emph{{Handbook of Perceptual Dialectology}}}.
\newblock John Benjamins Publishing Co.

\bibitem[{San et~al.(2021)San, Bartelds, Browne, Clifford, Gibson, Mansfield,
  Nash, Simpson, Turpin, Vollmer, Wilmoth, and Jurafsky}]{san2021leveraging}
Nay San, Martijn Bartelds, Mitchell Browne, Lily Clifford, Fiona Gibson, John
  Mansfield, David Nash, Jane Simpson, Myfany Turpin, Maria Vollmer, Sasha
  Wilmoth, and Dan Jurafsky. 2021.
\newblock \href {http://arxiv.org/abs/2103.14583} {Leveraging pre-trained
  representations to improve access to untranscribed speech from endangered
  languages}.

\bibitem[{Scharenborg(2007)}]{scharenborg2007}
Odette Scharenborg. 2007.
\newblock \href {https://doi.org/https://doi.org/10.1016/j.specom.2007.01.009}
  {{Reaching over the gap: A review of efforts to link human and automatic
  speech recognition research}}.
\newblock \emph{Speech Communication}, 49(5):336--347.
\newblock Bridging the Gap between Human and Automatic Speech Recognition.

\bibitem[{Schneider et~al.(2019)Schneider, Baevski, Collobert, and
  Auli}]{schneider2019wav2vec}
Steffen Schneider, Alexei Baevski, Ronan Collobert, and Michael Auli. 2019.
\newblock \href {https://doi.org/10.21437/Interspeech.2019-1873} {{wav2vec:
  Unsupervised Pre-Training for Speech Recognition}}.
\newblock In \emph{Proc. Interspeech 2019}, pages 3465--3469.

\bibitem[{Smith and Eisner(2005)}]{smith_contrastive_2005}
Noah~A. Smith and Jason Eisner. 2005.
\newblock \href {https://doi.org/10.3115/1219840.1219884} {Contrastive
  {{Estimation}}: {{Training Log}}-{{Linear Models}} on {{Unlabeled Data}}}.
\newblock In \emph{Proceedings of the 43rd {{Annual Meeting}} of the
  {{Association}} for {{Computational Linguistics}} ({{ACL}}'05)}, pages
  354--362, {Ann Arbor, Michigan}. {Association for Computational Linguistics}.

\bibitem[{Steiger(1980)}]{steiger1980tests}
James~H Steiger. 1980.
\newblock \href {https://doi.org/10.1037/0033-2909.87.2.245} {Tests for
  comparing elements of a correlation matrix.}
\newblock \emph{Psychological bulletin}, 87(2):245.

\bibitem[{Strycharczuk et~al.(2020)Strycharczuk, López-Ibáñez, Brown, and
  Leemann}]{Strycharczuk2020}
Patrycja Strycharczuk, Manuel López-Ibáñez, Georgina Brown, and Adrian
  Leemann. 2020.
\newblock \href {https://doi.org/10.3389/frai.2020.00048} {General northern
  english. exploring regional variation in the north of england with machine
  learning}.
\newblock \emph{Frontiers in Artificial Intelligence}, 3:48.

\bibitem[{Tenney et~al.(2019)Tenney, Das, and
  Pavlick}]{tenneyBERTRediscoversClassical2019}
Ian Tenney, Dipanjan Das, and Ellie Pavlick. 2019.
\newblock \href {https://doi.org/10.18653/v1/P19-1452} {{{BERT Rediscovers}}
  the {{Classical NLP Pipeline}}}.
\newblock In \emph{Proceedings of the 57th {{Annual Meeting}} of the
  {{Association}} for {{Computational Linguistics}}}, pages 4593--4601,
  {Florence, Italy}. {Association for Computational Linguistics}.

\bibitem[{Vaswani et~al.(2017)Vaswani, Shazeer, Parmar, Uszkoreit, Jones,
  Gomez, Kaiser, and Polosukhin}]{vaswaniAttentionAllYou2017}
Ashish Vaswani, Noam Shazeer, Niki Parmar, Jakob Uszkoreit, Llion Jones,
  Aidan~N Gomez, \L~ukasz Kaiser, and Illia Polosukhin. 2017.
\newblock \href
  {https://proceedings.neurips.cc/paper/2017/file/3f5ee243547dee91fbd053c1c4a845aa-Paper.pdf}
  {Attention is all you need}.
\newblock In \emph{Advances in Neural Information Processing Systems},
  volume~30. Curran Associates, Inc.

\bibitem[{Viglino et~al.(2019)Viglino, Motlicek, and Cernak}]{viglino2019end}
Thibault Viglino, Petr Motlicek, and Milos Cernak. 2019.
\newblock \href {https://doi.org/10.21437/Interspeech.2019-2122} {{End-to-End
  Accented Speech Recognition}}.
\newblock In \emph{Proc. Interspeech 2019}, pages 2140--2144.

\bibitem[{de~Vries et~al.(2020)de~Vries, van Cranenburgh, and
  Nissim}]{devriesWhatSpecialBERT2020}
Wietse de~Vries, Andreas van Cranenburgh, and Malvina Nissim. 2020.
\newblock \href {https://doi.org/10.18653/v1/2020.findings-emnlp.389} {What{'}s
  so special about {BERT}{'}s layers? a closer look at the {NLP} pipeline in
  monolingual and multilingual models}.
\newblock In \emph{Findings of the Association for Computational Linguistics:
  EMNLP 2020}, pages 4339--4350, Online. Association for Computational
  Linguistics.

\bibitem[{Weinberger and Kunath(2011)}]{weinberger2015speech}
Steven~H Weinberger and Stephen~A Kunath. 2011.
\newblock \href {https://doi.org/10.1163/9789401206884_014} {The speech accent
  archive: towards a typology of english accents}.
\newblock In \emph{Corpus-based studies in language use, language learning, and
  language documentation}, pages 265--281. Brill Rodopi.

\bibitem[{Wieling et~al.(2014{\natexlab{a}})Wieling, Bloem, Baayen, and
  Nerbonne}]{wieling2014determinants}
Martijn Wieling, Jelke Bloem, R~Harald Baayen, and John Nerbonne.
  2014{\natexlab{a}}.
\newblock \href {https://doi.org/http://dx.doi.org/10.15496/publikation-8628}
  {Determinants of english accents}.
\newblock \emph{Proceedings of the 6th Conference on Quantitative
  Investigations in Theoretical Linguistics}.

\bibitem[{Wieling et~al.(2014{\natexlab{b}})Wieling, Bloem, Mignella,
  Timmermeister, and Nerbonne}]{wieling2014a}
Martijn Wieling, Jelke Bloem, Kaitlin Mignella, Mona Timmermeister, and John
  Nerbonne. 2014{\natexlab{b}}.
\newblock \href {https://doi.org/https://doi.org/10.1163/22105832-00402001}
  {Measuring foreign accent strength in english: Validating levenshtein
  distance as a measure}.
\newblock \emph{Language Dynamics and Change}, 4(2):253 -- 269.

\bibitem[{Wieling et~al.(2007)Wieling, Heeringa, and
  Nerbonne}]{wieling2007aggregate}
Martijn Wieling, Wilbert Heeringa, and John Nerbonne. 2007.
\newblock \href
  {https://www.aup.nl/journal-downloads/taal-en-tongval/jrg_2007_vol_59_-_an_aggregate_analysis_of_pronunciation_in.pdf}
  {An aggregate analysis of pronunciation in the goeman-taeldeman-van
  reenen-project data}.
\newblock \emph{Taal en Tongval}, 59(1):84--116.

\bibitem[{Wieling et~al.(2012)Wieling, Margaretha, and
  Nerbonne}]{wieling2012inducing}
Martijn Wieling, Eliza Margaretha, and John Nerbonne. 2012.
\newblock \href {https://doi.org/https://doi.org/10.1016/j.wocn.2011.12.004}
  {Inducing a measure of phonetic similarity from pronunciation variation}.
\newblock \emph{Journal of Phonetics}, 40(2):307--314.

\bibitem[{Wieling and Nerbonne(2015)}]{wieling2015}
Martijn Wieling and John Nerbonne. 2015.
\newblock \href {https://doi.org/10.1146/annurev-linguist-030514-124930}
  {{Advances in Dialectometry}}.
\newblock \emph{Annual Review of Linguistics}, 1(1):243--264.

\bibitem[{Wieling et~al.(2011)Wieling, Nerbonne, and Baayen}]{wieling2011}
Martijn Wieling, John Nerbonne, and R.~Harald Baayen. 2011.
\newblock \href {https://doi.org/10.1371/journal.pone.0023613} {Quantitative
  social dialectology: Explaining linguistic variation geographically and
  socially}.
\newblock \emph{PLOS ONE}, 6(9):1--14.

\bibitem[{Yuan and Liberman(2008)}]{yuan2008speaker}
Jiahong Yuan and Mark Liberman. 2008.
\newblock \href {https://doi.org/10.1121/1.2935783} {Speaker identification on
  the scotus corpus}.
\newblock \emph{Journal of the Acoustical Society of America}, 123(5):3878.

\bibitem[{Zhao and Wang(2013)}]{zhao2013}
Xiaojia Zhao and DeLiang Wang. 2013.
\newblock \href {https://doi.org/10.1109/ICASSP.2013.6639061} {Analyzing noise
  robustness of mfcc and gfcc features in speaker identification}.
\newblock In \emph{2013 IEEE International Conference on Acoustics, Speech and
  Signal Processing}, pages 7204--7208.

\bibitem[{Żelasko et~al.(2020)Żelasko, Moro-Velázquez, Hasegawa-Johnson,
  Scharenborg, and Dehak}]{zelasko20}
Piotr Żelasko, Laureano Moro-Velázquez, Mark Hasegawa-Johnson, Odette
  Scharenborg, and Najim Dehak. 2020.
\newblock \href {https://doi.org/10.21437/Interspeech.2020-2513} {{That Sounds
  Familiar: An Analysis of Phonetic Representations Transfer Across
  Languages}}.
\newblock In \emph{Proc. Interspeech 2020}, pages 3705--3709.

\end{thebibliography}

\appendix
\section{Appendices}
\label{sec:appendix}
Appendix~\ref{appendix:models} provides all relevant technical details about the neural models used in this paper. Appendix~\ref{appendix:layers} visualizes the performance per layer for each of the neural models applied to different datasets. 
\subsection{Technical details neural models}
\label{appendix:models}

\subsubsection{wav2vec}
\texttt{wav2vec} (\texttt{w2v}) is a self-supervised pre-trained neural model that has been developed for speech recognition \citep{schneider2019wav2vec}.
This model consists of an encoder network and an aggregator network, and is trained in two model configurations, namely small and large.
In this paper, we include the large model configuration to compute acoustic pronunciation distances, because the small model configuration is only trained on a subset of the Librispeech dataset, whereas the large model configuration uses the full Librispeech dataset.

The encoder network of the large model configuration consists of seven convolutional layers that create a dense representation of audio with a sliding window strategy (stride is 10ms, window size is 30ms).
The dense output representations are aggregated by the aggregator network with 12 convolutional layers.
The output of the encoder network is based on 30ms of audio in steps of 10ms, whereas the output of the aggregator network is based on windows of 810ms.

\texttt{w2v} is trained to predict upcoming audio frames of a speech utterance based on preceding frames.
Inspired by \texttt{word2vec} \citep{mikolov_distributed_2013}, the model is trained with a contrastive loss objective, which is defined as the probability of distinguishing the actual frame from ten negative example frames sampled from the same utterance.
To this end, \texttt{w2v} should be sensitive to content in the actual target frame.
A regular loss objective for target frame prediction would let the model learn to replicate features that are consistent within their context, such as voice properties and noise.
However, these features are not only undesirable, but they are also likely to be present as negative evidence in random negative samples from the same fragment.
Therefore, a contrastive loss objective is more likely to reach better performance
\citep{smith_contrastive_2005}.

During inference, speech features can be extracted from the encoder (512 dimensions) or the aggregator (512 dimensions).
The encoder represents features within a 30ms context window, whereas the aggregator outputs input reconstructions based on 810ms of context. We select the features from the encoder, as initial experiments showed that this resulted in the highest performance. 

\subsubsection{vq-wav2vec + BERT}
\texttt{vq-wav2vec} is an extension of \texttt{w2v} with the same architecture, except for the addition of a quantization layer between the encoder and the aggregator networks \citep{baevski2019vq}.
This quantization layer creates a discrete representation of the dense encoder outputs.
Quantization is done with either the Gumbel Softmax differentiable argmax approach \citep{jang_categorical_2017}, or with online K-means clustering \citep{van_den_oord_neural_2017}.

Discretization of \texttt{w2v} enables the use of algorithms that require discrete input, such as \texttt{BERT}.
\texttt{BERT} is a non-recurrent neural network architecture, and training method, that can process sequential data \citep{devlin_bert_2019}. 
Traditional neural methods require iterative processing of sequential data (i.e.~recurrent neural networks), but the self-attention mechanism in the Transformer layers of \texttt{BERT} ensures that entire sequences can be processed at once \citep{vaswaniAttentionAllYou2017}. 
The self-attention mechanism works like a weighting mechanism for context in a sequence, and is based on content and position. 
Context-based representations are therefore only influenced by close context that is likely to be informative.
The \texttt{BERT} model has shown to be highly scalable for text processing \citep{devlin_bert_2019}.

\citet{baevski2019vq} applied a 12 layer \texttt{BERT} model to the discrete output of \texttt{vq-wav2vec}.
The \texttt{BERT} model is trained by masking random spans of 100ms of audio that have to be predicted as a pre-training objective, where each frame has a five percent chance of starting a masked sequence.
The \texttt{vq-wav2vec} algorithm with the \texttt{BERT} extension was found to outperform the regular \texttt{vq-wav2vec} model on speech recognition tasks.
Therefore, the \texttt{BERT} model may learn representations of speech that differentiate sounds in speech more clearly than the \texttt{vq-wav2vec} model itself. 
Moreover, \citet{baevski2019vq} show that, when applied to speech recognition, the full pipeline is slightly more effective with K-means quantization than Gumbel softmax quantization. We therefore use the \texttt{vq-wav2vec} algorithm with the \texttt{BERT} extension. In the following, we refer to this variant as \texttt{vqw2v}.

Speech representations can be extracted from multiple layers in the \texttt{vqw2v} pipeline.
The \texttt{vqw2v} model itself can yield representations after encoding (512 dimensions), quantization (512 dimensions), and aggregation (512 dimensions).
Additionally, the separately trained \texttt{BERT} model can provide representations after each of the 12 Transformer layers (768 dimensions).
These Transformer layers can iteratively make embeddings more informative, but the final layers do not tend to be the most informative layers for downstream tasks \citep{tenneyBERTRediscoversClassical2019, devriesWhatSpecialBERT2020}.
A likely explanation is that informative abstractions and generalizations in hidden layers are discarded in favor of actual target output.
We choose the best-performing Transformer layer for our task based on a validation set from each dataset.

\subsubsection{wav2vec 2.0}
For the \texttt{vqw2v} algorithm, the Transformer layers in the \texttt{BERT} model are trained as a separate step after training \texttt{vqw2v}.
For \texttt{w2v2}, the convolutional aggregator in \texttt{vqw2v} is replaced by a Transformer network \citep{baevski2020wav2vec}.
This has led to improved performance in speech recognition compared to \texttt{vqw2v} with \texttt{BERT}, suggesting that \texttt{w2v2} may contain better speech representations.
Unlike \texttt{vqw2v} with the \texttt{BERT} extension, \texttt{w2v2} is trained as a single end-to-end model, and therefore the encoder outputs are optimized for use in the Transformer.
The final pipeline of \texttt{w2v2} consists of a convolutional encoder, a quantizer, and a Transformer model.
Gumbel softmax quantization is used in \texttt{w2v2}, and the best-performing variant of \texttt{w2v2} in speech recognition contains a fixed amount of 24 Transformer layers.  

Whereas the original \texttt{w2v} aggregator is trained to predict speech frames based on the preceding frames, the \texttt{w2v2} Transformer aggregator has to predict spans of randomly masked frames with the full fragment as context.
The task of predicting single random frames is considered to be trivial, and therefore sequences of 10 consecutive frames are masked at each randomly sampled position with a probability of 6.5\% for each frame to start a masked sequence.
Effectively, during pre-training, 49\% of all frames are masked in blocks with an average duration of 299ms.
Similarly to \texttt{w2v}, the \texttt{w2v2} model is trained with a contrastive loss function based on negative sampling.

Pre-trained models can be fine-tuned for speech recognition using labeled data.
The models are augmented by adding a randomly initialized linear projection to the Transformer network.
This projection contains an amount of classes that is similar to the size of the vocabulary of the labeled data.
Using connectionist temporal classification \citep{graves2006connectionist}, probability scores of a textual output sequence can be obtained based on the vocabulary of the task.

Similar to \texttt{vqw2v}, embeddings can be extracted from the encoder (512 dimensions), the quantizer (768 dimensions), or the (fine-tuned) Transformer layers (1024 dimensions).
We investigate the monolingual English \texttt{w2v2} model pre-trained on LS960, and a version that has subsequently been fine-tuned on 960 hours on labeled data from Librispeech (denoted by \texttt{w2v2-en}).
These models are chosen because they are the largest models available, and \citet{baevski2020wav2vec} showed that increasing the model size improved performance on all evaluated speech recognition tasks.
We select the best-performing Transformer layer based on a validation set from each dataset.

\subsubsection{XLSR}
\texttt{XLSR} builds on \texttt{w2v2} by extending pre-training to 56,000 hours of speech from a total of 53 languages \citep{conneau2020unsupervised}.
These 53 languages are obtained from the Common Voice dataset (which contains read speech from 36 European languages; \citealp{ardila2019common}), the BABEL dataset (conversational telephone speech from 17 Asian and African languages; \citealp{gales2014speech}), and Multilingual Librispeech (audio books from 8 European languages; \citealp{pratap2020mls}).
Note that the majority of the pre-training data consists of English speech from Multilingual Librispeech dataset (44,000 hours), and that some languages occur in more than one dataset. Consequently, the total number of languages included during pre-training is 53.
Due to the multilingual setup of the \texttt{XLSR} model, we expect improved task performance when using our Norwegian dataset, compared to the monolingual models.

The architecture of \texttt{XLSR} is similar to \texttt{w2v2}, with the exception that a single set of discrete speech representations is learned by the quantizer on the basis of the encoder output.
The discrete representations are subsequently shared across languages, creating connections between the different pre-training languages. 

Similar to \texttt{w2v2}, embeddings can be extracted from the encoder (512 dimensions), the quantizer (768 dimensions), or the (fine-tuned) Transformer layers (1024 dimensions).
We use the multilingual \texttt{XLSR} model pre-trained on 53 languages, and fine-tuned on languages from the Common Voice dataset (version 6.1) \citep{ardila2019common}.
Specifically, we consider \texttt{XLSR} fine-tuned on English (1,686 hours, denoted by \texttt{XLSR-en}) and Swedish (12 hours, denoted by \texttt{XLSR-sv}), as they match (or, in the case of Swedish, is most similar to) the languages in our evaluation datasets. As before, we select the best-performing Transformer layer based on a validation set from each dataset.

\subsubsection{DeCoAR} 
The \texttt{w2v} model uses convolutional layers to create representations of audio based on close context, whereas newer Transformer-based models use the entire audio fragment as context.
\texttt{DeCoAR} uses an alternative method to process the audio sequences.
Before \texttt{BERT} models were used in natural language processing, language models relied on recurrent neural models that process items in a sequence, one step at a time.
Representations of each item are, in this case, based on the preceding representation.
The most commonly used model in natural language processing that uses this method is \texttt{ELMo} \citep{petersDeepContextualizedWord2018}.
This model uses a stacked LSTM network for creating contextualized word embeddings.

\citet{ling2020deep} apply the bi-directional LSTM method, that was proposed by \citet{petersDeepContextualizedWord2018}, to encode acoustic speech signals.
The resulting \texttt{DeCoAR} model takes 40-dimensional log filterbank features as its input, and is trained to reconstruct the same features as its output.
A filterbank transformation subsequently extracts frequency bands by dividing the frequency range into 40 triangular overlapping bins.
These features are extracted with a 25ms sliding window and a stride of 10ms.
\texttt{DeCoAR} consists of four bi-directional LSTM layers, each having 1024 cells.
The output representation of \texttt{DeCoAR} is the concatenation of the forward and backward directions, and therefore consists of 2048 dimensions.

The novel \texttt{DeCoAR} model was shown to outperform \texttt{w2v} on a set of tasks, including phone classification \citep{ma:21}.
% Representations extracted from \texttt{DeCoAR} were shown to outperform \texttt{w2v} on a set of tasks, including phone classification and vowel detection \citep{ma:21}.
Therefore, architectural differences of \texttt{DeCoAR} with \texttt{w2v}-based models may show performance differences when applied to other tasks, such as modeling speech variation.
\newpage

\onecolumn
\subsection{Layer performance}
\label{appendix:layers}

\begin{figure*}[ht]
    %  \centering
      \begin{subfigure}[b]{0.49\textwidth}
         \centering
         \includegraphics[width=2.2in]{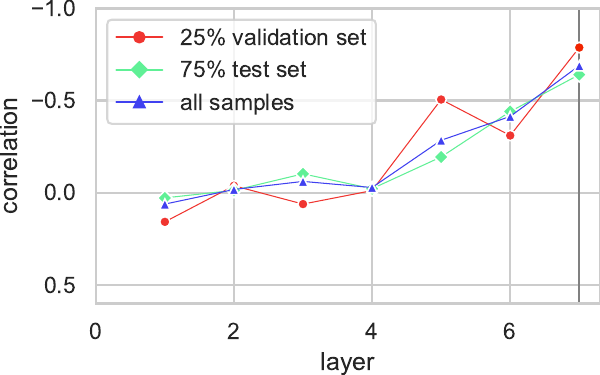}
         \caption{Pre-trained monolingual English \texttt{w2v}}
         \label{fig:curveW2VSAA}
      \end{subfigure}
      \begin{subfigure}[b]{0.49\textwidth}
         \centering
         \includegraphics[width=2.2in]{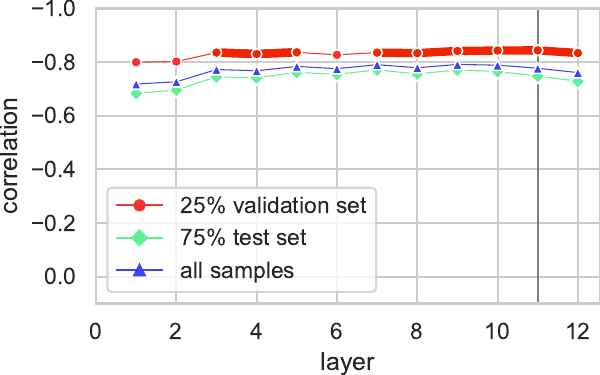}
         \caption{Pre-trained monolingual English \texttt{vqw2v}}
         \label{fig:curveVQW2VSAA}
     \end{subfigure}
     \par\smallskip
     \begin{subfigure}[b]{0.49\textwidth}
         \centering
         \includegraphics[width=2.2in]{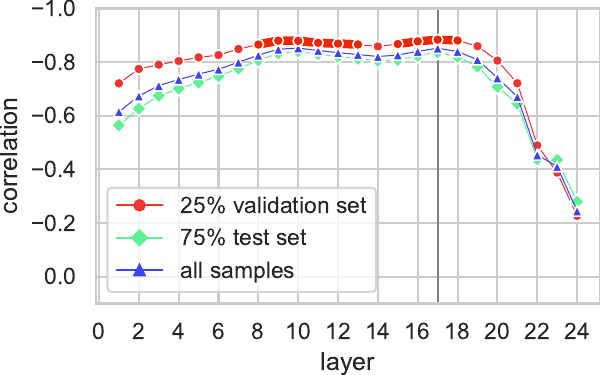}
         \caption{Pre-trained monolingual English \texttt{w2v2}}
         \label{fig:curveW2V2SAA}
     \end{subfigure}
     \begin{subfigure}[b]{0.49\textwidth}
         \centering
         \includegraphics[width=2.2in]{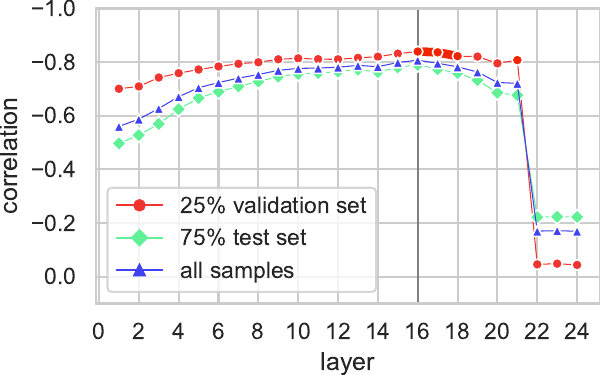}
         \caption{Pre-trained multilingual \texttt{XLSR}}
         \label{fig:curveXLSRSAA}
     \end{subfigure}
     \par\smallskip
     \begin{subfigure}[b]{0.49\textwidth}
         \centering
         \includegraphics[width=2.2in]{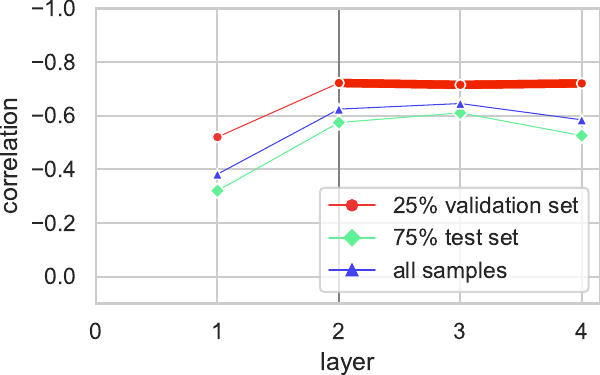}
         \caption{Pre-trained monolingual English \texttt{DeCoAR}}
         \label{fig:curveDECOARSAA}
     \end{subfigure}
     \begin{subfigure}[b]{0.49\textwidth}
         \centering
         \includegraphics[width=2.2in]{figures/SAA/wav2vec2-large-960h.pdf}
         \caption{Fine-tuned monolingual \texttt{w2v2-en}}
         \label{fig:curveW2V2-ENSAA}
      \end{subfigure}
      \par\smallskip
      \begin{subfigure}[b]{0.49\textwidth}
         \centering
         \includegraphics[width=2.2in]{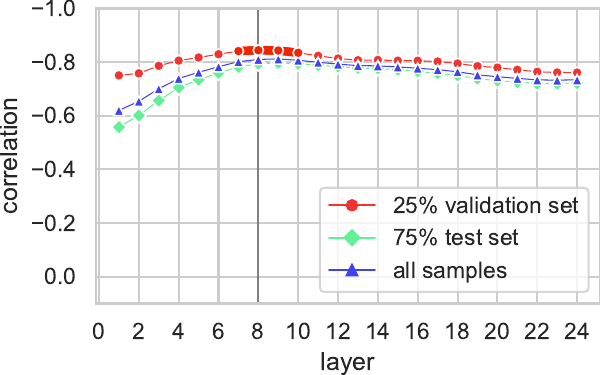}
         \caption{Fine-tuned multilingual \texttt{XLSR-en}}
         \label{fig:curveXLSR-ENSAA}
     \end{subfigure}
     \caption{Pearson correlation coefficients of acoustic distances compared to human accent ratings for different layers in \texttt{w2v}, \texttt{vqw2v}, \texttt{DeCoAR}, and the \texttt{w2v2} and \texttt{XLSR} models. The vertical line marks the layer that was chosen as the best-performing layer based on the 25\% validation set of the Speech Accent Archive dataset. Layers with a correlation that is not significantly different ($p > 0.05$) from the optimal layer are indicated by the thick red line.}
     \label{fig:curveSAA}
\end{figure*}

\begin{figure*}[ht]
    %  \centering
       \begin{subfigure}[b]{0.49\textwidth}
         \centering
         \includegraphics[width=2.2in]{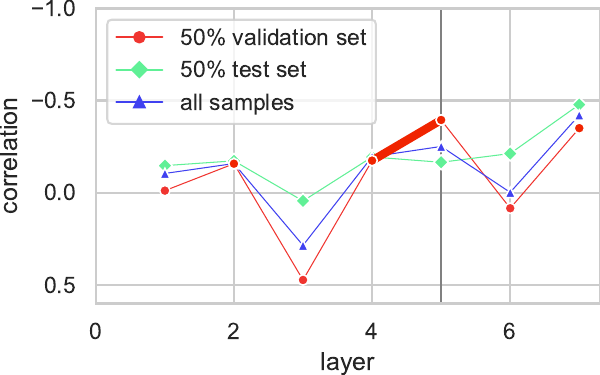}
         \caption{Pre-trained monolingual English \texttt{w2v}}
         \label{fig:curveW2VDSD}
      \end{subfigure}
      \begin{subfigure}[b]{0.49\textwidth}
         \centering
         \includegraphics[width=2.2in]{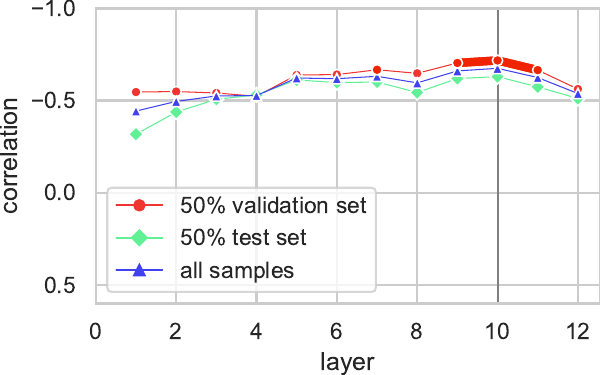}
         \caption{Pre-trained monolingual English \texttt{vqw2v}}
         \label{fig:curveVQW2VDSD}
     \end{subfigure}
     \par\smallskip
     \begin{subfigure}[b]{0.49\textwidth}
         \centering
         \includegraphics[width=2.2in]{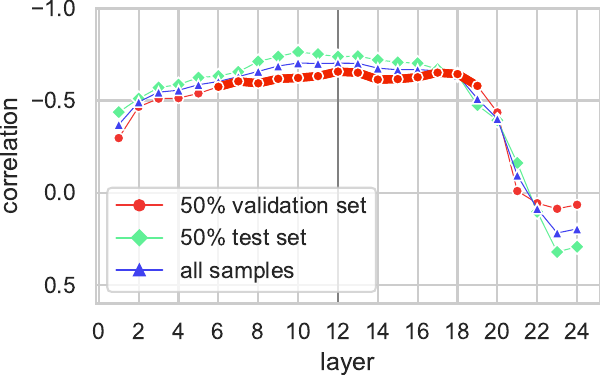}
         \caption{Pre-trained monolingual English \texttt{w2v2}}
         \label{fig:curveW2V2DSD}
     \end{subfigure}
     \begin{subfigure}[b]{0.49\textwidth}
         \centering
         \includegraphics[width=2.2in]{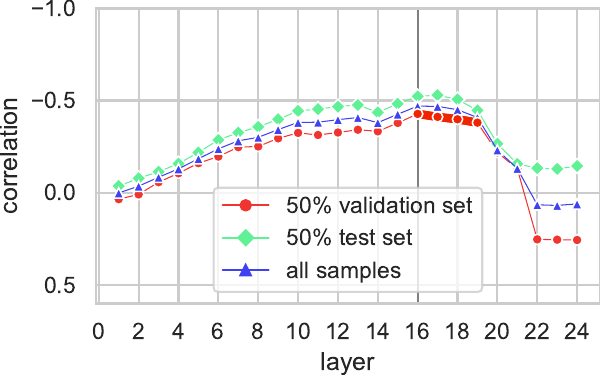}
         \caption{Pre-trained multilingual \texttt{XLSR}}
         \label{fig:curveXLSRDSD}
     \end{subfigure}
     \par\smallskip
     \begin{subfigure}[b]{0.49\textwidth}
         \centering
         \includegraphics[width=2.2in]{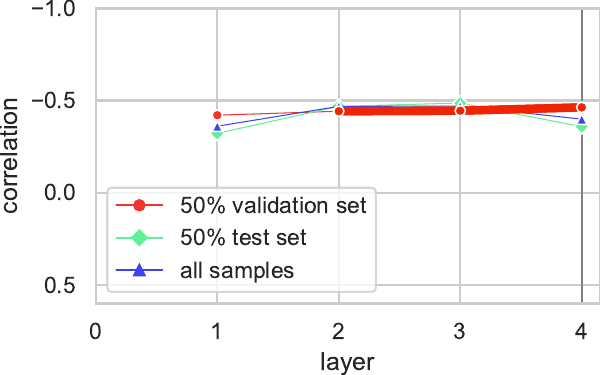}
         \caption{Pre-trained monolingual English \texttt{DeCoAR}}
         \label{fig:curveDECOARDSD}
     \end{subfigure}
     \begin{subfigure}[b]{0.49\textwidth}
         \centering
         \includegraphics[width=2.2in]{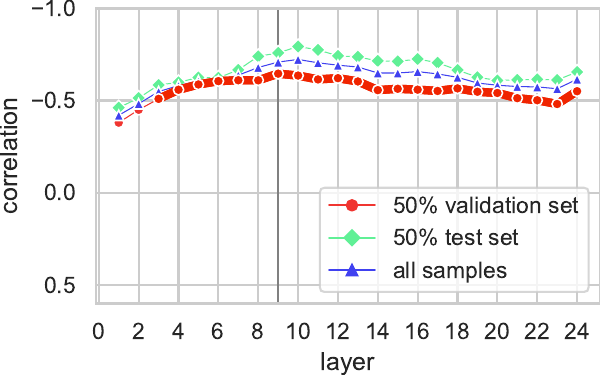}
         \caption{Fine-tuned monolingual \texttt{w2v2-en}}
         \label{fig:curveW2V2-ENDSD}
      \end{subfigure}
      \par\smallskip
      \begin{subfigure}[b]{0.49\textwidth}
         \centering
         \includegraphics[width=2.2in]{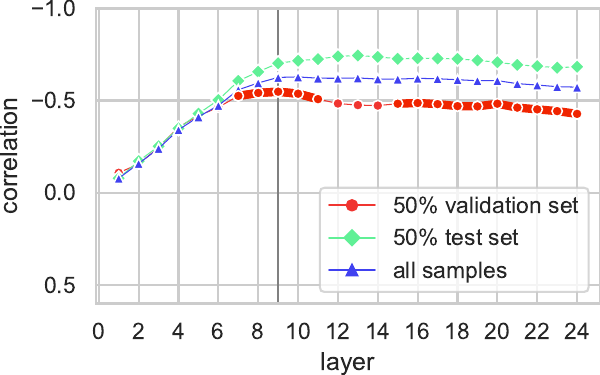}
         \caption{Fine-tuned multilingual \texttt{XLSR-en}}
         \label{fig:curveXLSR-ENDSD}
     \end{subfigure}
     \caption{Pearson correlation coefficients of acoustic distances compared to human accent ratings for different layers in \texttt{w2v}, \texttt{vqw2v}, \texttt{DeCoAR}, and the \texttt{w2v2} and \texttt{XLSR} models. The vertical line marks the layer that was chosen as the best-performing layer based on the 50\% validation set of the Dutch speakers dataset. Layers with a correlation that is  not significantly different ($p > 0.05$) from the optimal layer are indicated by the thick red line.}
     \label{fig:curveDSD}
\end{figure*}

\begin{figure*}[ht]
     \centering
    %  \begin{subfigure}[b]{0.49\textwidth}
    %      \centering
    %      \includegraphics[width=3in]{figures/Norwegian/wav2vec2-large.pdf}
    %      \caption{Pre-trained monolingual English \texttt{w2v2}.}
    %      \label{fig:curveW2V2NOS}
    %  \end{subfigure}
    %  \hfill
    %  \begin{subfigure}[b]{0.49\textwidth}
    %      \centering
    %      \includegraphics[width=3in]{figures/Norwegian/wav2vec2-large-xlsr-53.pdf}
    %      \caption{Pre-trained multilingual \texttt{XLSR}.}
    %      \label{fig:curveXLSRNOS}
    %  \end{subfigure}
     \begin{subfigure}[b]{0.49\textwidth}
         \centering
         \includegraphics[width=2.2in]{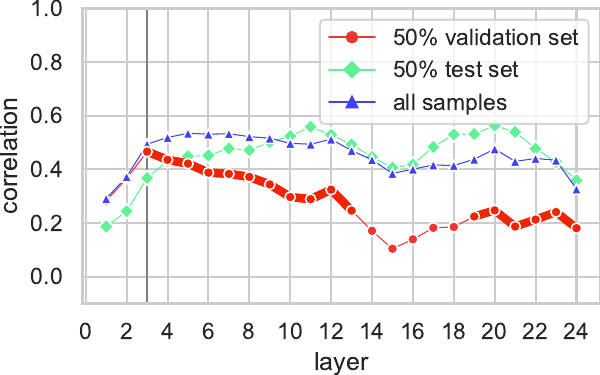}
         \caption{Fine-tuned monolingual \texttt{w2v2-en}}
         \label{fig:curveW2V2-ENNOS}
      \end{subfigure}
      \begin{subfigure}[b]{0.49\textwidth}
         \centering
         \includegraphics[width=2.2in]{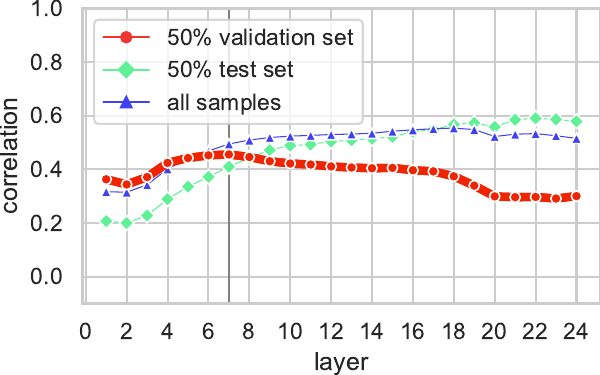}
         \caption{Fine-tuned multilingual \texttt{XLSR-sv}}
         \label{fig:curveXLSR-SVNOS}
     \end{subfigure}
     \caption{Pearson correlation coefficients of acoustic distances compared to human accent ratings for different Transformer layers in the \texttt{w2v2-en} and \texttt{XLSR-sv} models. The vertical line marks the layer that was chosen as the best-performing layer based on the 50\% validation set of the Norwegian dialects dataset. Layers with a correlation that is  not significantly different ($p > 0.05$) from the optimal layer are indicated by the thick red line.}
     \label{fig:curveNOS}
\end{figure*}

\end{document}